\documentclass{article}

\usepackage{pgfplots}

\pgfplotsset{compat=1.17}

\usetikzlibrary{spy,patterns}
\usepackage{colortbl}
\usepackage{graphicx}
\usepackage[Export]{adjustbox}
\usepackage{tabularx}
\usepackage{cellspace}
\usepackage{amsmath}
\usepackage{amssymb}
\usepackage{mathtools}
\usepackage{amsthm}
\usepackage{xspace}
\usepackage{multirow}
\usepackage{makecell}
\usepackage{enumitem}

\usepackage{microtype}
\usepackage{graphicx}
\usepackage{subfigure}
\usepackage{booktabs} 
\usepackage{hyperref}

\usepackage[accepted]{icml2025}

\usepackage{amsmath}
\usepackage{amssymb}
\usepackage{mathtools}
\usepackage{amsthm}
\usepackage{xspace}
\usepackage{caption}
\usepackage[capitalize,noabbrev]{cleveref}
\theoremstyle{plain}

\theoremstyle{definition}

\theoremstyle{remark}

\usepackage[textsize=tiny]{todonotes}
\usepackage{booktabs}
\usepackage{colortbl}

\pgfplotsset{
  colormap={myfade}{
    rgb255(0cm)=(44,62,80)
    rgb255(1cm)=(52,152,219)
  },
}

\icmltitlerunning{Equivariance Regularized Latent Space for Improved Generative Image Modeling}

\usetikzlibrary{arrows.meta,shapes,calc,matrix,fit,positioning,backgrounds,decorations.markings}
\pgfplotsset{compat=1.9}
\usepackage{xstring}

\usepgfplotslibrary{external}

\IfBeginWith*{\jobname}{fig/extern/}{\finalcopy}{}

\tikzstyle{every picture}+=[
	remember picture,
	every text node part/.style={align=center},
	every matrix/.append style={ampersand replacement=\&},
]
\tikzstyle{tight} = [inner sep=0pt,outer sep=0pt]
\tikzstyle{node}  = [draw,circle,tight,minimum size=12pt,anchor=center]
\tikzstyle{op}    = [draw,circle,tight]
\tikzstyle{dot}   = [fill,draw,circle,inner sep=1pt,outer sep=0]
\tikzstyle{pt}    = [fill,draw,circle,inner sep=1.5pt,outer sep=.2pt]
\tikzstyle{box}   = [draw,rectangle,inner sep=3pt]
\tikzstyle{high}  = [black!60]
\tikzstyle{group} = [high,box,opacity=.5]
\tikzstyle{dim1}  = [fill opacity=.3,text opacity=1]
\tikzstyle{dim2}  = [fill opacity=.5,text opacity=1]
\tikzstyle{dim3}  = [fill opacity=.7,text opacity=1]
\tikzstyle{rectc} = [tight,transform shape]
\tikzstyle{rect}  = [rectc,anchor=south west]

\tikzset{every mark/.append style={solid}}
\pgfplotsset{
	grid=both, width=\columnwidth, try min ticks=5,
	every axis/.append style={font=\small},
	every axis plot/.append style={thick,mark=none,mark size=1.8,tension=0.18},
	legend cell align=left, legend style={fill opacity=0.8},
	xticklabel={\pgfmathprintnumber[assume math mode=true]{\tick}},
	yticklabel={\pgfmathprintnumber[assume math mode=true]{\tick}},
	nodes near coords math/.style={
		nodes near coords={\pgfmathprintnumber[assume math mode=true]{\pgfplotspointmeta}},
	},
}

\pgfplotsset{
	dash/.style={mark=o,dashed,opacity=0.6},
	dott/.style={mark=o,dotted,opacity=0.6},
	nolim/.style={enlargelimits=false},
	plain/.style={every axis plot/.append style={},nolim,grid=none},
}

\pgfdeclarelayer{bg4}
\pgfdeclarelayer{bg3}
\pgfdeclarelayer{bg2}
\pgfdeclarelayer{bg1}
\pgfdeclarelayer{fg1}
\pgfdeclarelayer{fg2}
\pgfdeclarelayer{fg3}
\pgfdeclarelayer{fg4}
\pgfsetlayers{bg4,bg3,bg2,bg1,main,fg1,fg2,fg3,fg4}

\pgfkeys{/tikz/.cd, aspect/.store in=\aspect, aspect=1}
\pgfkeys{/tikz/.cd, depth/.store in=\depth, depth=.5}
\pgfkeys{/tikz/.cd, stepx/.store in=\step, stepx=1}

\tikzstyle{geom} = [line join=bevel,aspect=1,depth=.5,z={(\depth*\aspect,\depth)}]
\tikzstyle{wire} = [geom,draw,thick]

\def\cx[#1,#2,#3]{#1}
\def\cy[#1,#2,#3]{#2}
\def\cz[#1,#2,#3]{#3}
\def\ex[#1,#2,#3]{#1,0,0}
\def\ey[#1,#2,#3]{0,#2,0}
\def\ez[#1,#2,#3]{0,0,#3}

\newcommand{\Th}[1]{\textsc{#1}}

\newcommand{\mc}[2]{\multicolumn{#1}{c}{#2}}

\newcommand{\red}[1]{{\textcolor{red}{#1}}}

\newcommand{\citeme}[1]{\red{[XX]}}
\newcommand{\refme}[1]{\red{(XX)}}

\newcommand{\floor}[1]{\left\lfloor{#1}\right\rfloor}

\makeatletter
\newcommand*\bdot{\mathpalette\bdot@{.7}}
\newcommand*\bdot@[2]{\mathbin{\vcenter{\hbox{\scalebox{#2}{$\m@th#1\bullet$}}}}}
\makeatother

\makeatletter
\DeclareRobustCommand\onedot{\futurelet\@let@token\@onedot}
\def\@onedot{\ifx\@let@token.\else.\null\fi\xspace}

\makeatother

\newcommand{\dit}{\texttt{DiT}\xspace}
\newcommand{\sit}{\texttt{SiT}\xspace}
\newcommand{\ditbtwo}{\texttt{DiT-B/2}\xspace}
\newcommand{\sitbtwo}{\texttt{SiT-B/2}\xspace}

\newcommand{\ditxltwo}{\texttt{DiT-XL/2}\xspace}

\newcommand{\sitxltwo}{\texttt{SiT-XL/2}\xspace}
\newcommand{\sdvae}{\texttt{SD-VAE}\xspace}
\newcommand{\sdxlvae}{\texttt{SDXL-VAE}\xspace}

\newcommand{\maskgit}{\texttt{MaskGIT}\xspace}

\newcommand{\our}{\texttt{EQ-VAE}\xspace}

\definecolor{ForestGreen}{RGB}{34,139,34}
\definecolor{Orange}{RGB}{1.0, 0.55, 0.0}
\definecolor{Purple}{RGB}{0.58, 0.44, 0.86}
\definecolor{softblue}{rgb}{0.2, 0.4, 0.8} 
\definecolor{teal}{rgb}{0.1, 0.6, 0.6} 
\definecolor{forestgreen}{rgb}{0.1, 0.6, 0.2} 

\definecolor{vibrantorange}{rgb}{1.0, 0.5, 0.0}

\definecolor{deepblue}{rgb}{0.0, 0.0, 0.8}

\definecolor{TableColor}{rgb}{0.92, 0.95, 0.99}  

\definecolor{DarkenedMagenta}{RGB}{225,0,100}

\newcommand{\grayhline}{\arrayrulecolor[HTML]{999999}\hline\arrayrulecolor{black}} 

\newcommand{\graycline}[1]{%
  \arrayrulecolor[HTML]{999999}%
  \cline{#1}%
  \arrayrulecolor{black}%
}

\newcommand{\Eq}[1]{\hyperref[#1]{Eq.~(\ref{#1})}}
\newcommand{\Equation}[1]{\hyperref[#1]{Equation~(\ref{#1})}}
\newcommand{\secref}[1]{\hyperref[#1]{Sec.~\ref{#1}}}

\begin{document}

\twocolumn[
\icmltitle{\textbf{\texttt{EQ-VAE}}: Equivariance Regularized Latent Space for\\Improved Generative Image Modeling}

\icmlsetsymbol{equal}{*}

\begin{icmlauthorlist}
\icmlauthor{Theodoros Kouzelis}{arch,ntua}
\icmlauthor{Ioannis Kakogeorgiou}{arch}
\icmlauthor{Spyros Gidaris}{valeo}
\icmlauthor{Nikos Komodakis}{arch,uoc,forth}

\end{icmlauthorlist}

\icmlaffiliation{arch}{Archimedes,Athena Reaserch Center, Greece}
\icmlaffiliation{valeo}{valeo.ai, France}
\icmlaffiliation{ntua}{National Technical University of Athens, Greece}
\icmlaffiliation{uoc}{University of Crete, Greece}
\icmlaffiliation{forth}{IACM-Forth, Greece}

\icmlcorrespondingauthor{Theodoros Kouzelis}{theodoros.kouzelis@athenarc.gr}

\icmlkeywords{Machine Learning, ICML}

\vskip 0.3in

]

\printAffiliationsAndNotice{}

\addtocontents{toc}{\protect\setcounter{tocdepth}{-1}}

\begin{abstract}
Latent generative models have emerged as a leading approach for high-quality image synthesis. These models rely on an autoencoder to compress images into a latent space, followed by a generative model to learn the latent distribution. 
We identify that existing autoencoders lack equivariance to semantic-preserving transformations like scaling and rotation, resulting in complex latent spaces that hinder generative performance. To address this, we propose \our, a simple regularization approach that enforces equivariance in the latent space, reducing its complexity without degrading reconstruction quality. By fine-tuning pre-trained autoencoders with \our, we enhance the performance of several state-of-the-art generative models, including \dit, \sit, \texttt{REPA} and \maskgit, achieving a $\times 7$ speedup on \ditxltwo with only five epochs of \sdvae fine-tuning. \our is compatible with both continuous and discrete autoencoders, thus offering a versatile enhancement for a wide range of latent generative models. Project page and code: \url{https://eq-vae.github.io/}.
\end{abstract}

\section{Introduction}
\label{sec:intro}

\begin{figure*}[tb]
\centering

\begin{minipage}[t]{0.37\textwidth}
  \vspace{0pt}%
  \centering
  \label{fig:teaser_latent}
  \footnotesize

  \newcommand{\myfigA}[1]{\includegraphics[width=0.258\textwidth,valign=c]{#1}}

  \setlength{\tabcolsep}{1pt}

\begin{tabular}{@{}c@{\hspace{2pt}}|@{\hspace{2pt}}cc@{\hspace{2pt}}|@{\hspace{2pt}}cc@{}}
    {\small \texttt{\textbf{Image}}} &
    {\small  \texttt{\textbf{SD-VAE}}} &
    {\small \makecell{\texttt{\textbf{+Ours}}}} &
    {\small \texttt{\textbf{SDXL-VAE}}} &
    {\small \makecell{ \texttt{\textbf{+Ours}}}} \\

    \myfigA{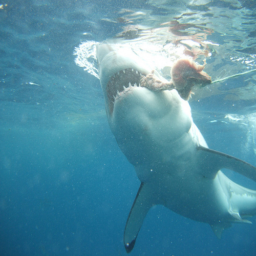} &
    \myfigA{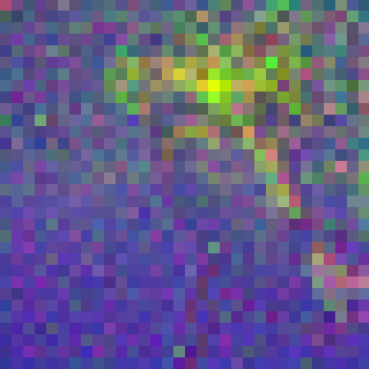} &
    \myfigA{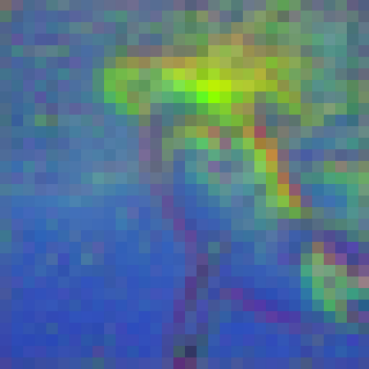} &
    \myfigA{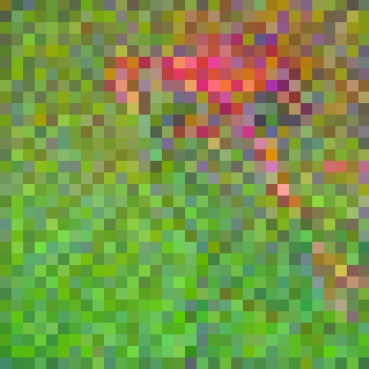} &
    \myfigA{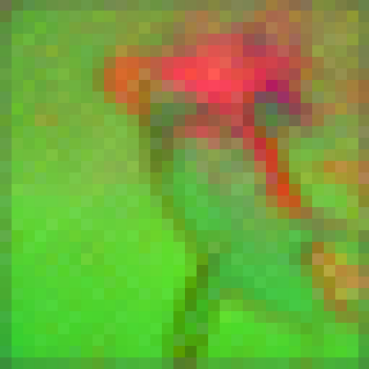} \\

    \multicolumn{5}{c}{\vspace{-2.1ex}}\\

    \myfigA{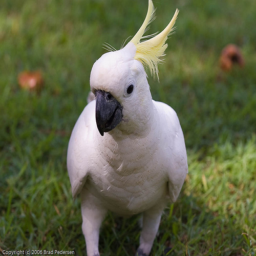} &
    \myfigA{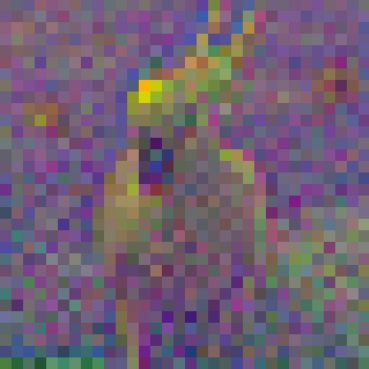} &
    \myfigA{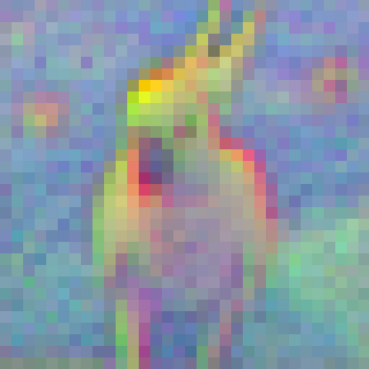} &
    \myfigA{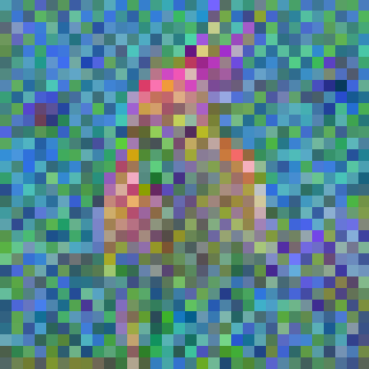} &
    \myfigA{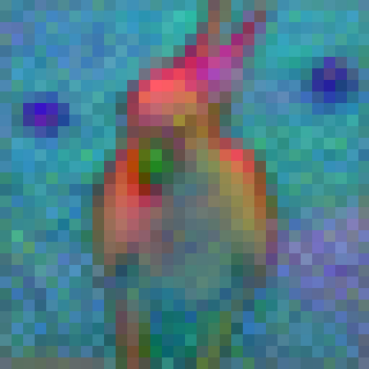} \\

    \multicolumn{5}{c}{\vspace{-2.1ex}}\\

    \myfigA{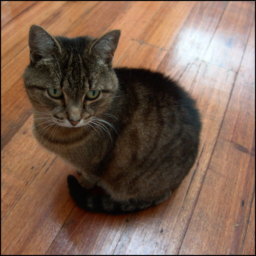} &
    \myfigA{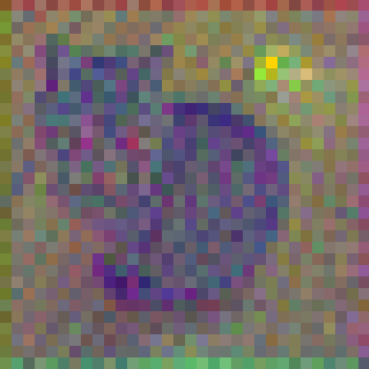} &
    \myfigA{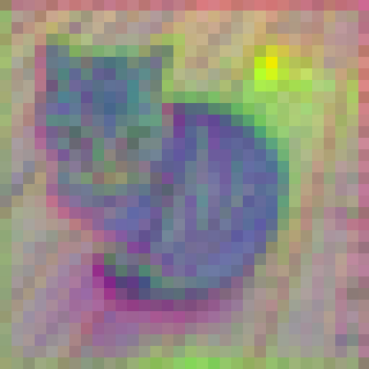} &
    \myfigA{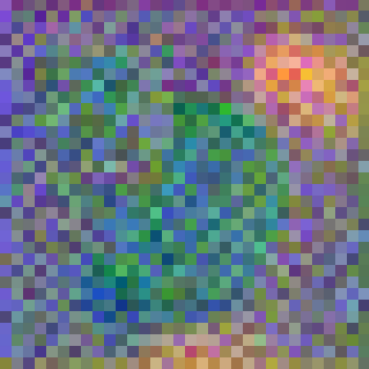} &
    \myfigA{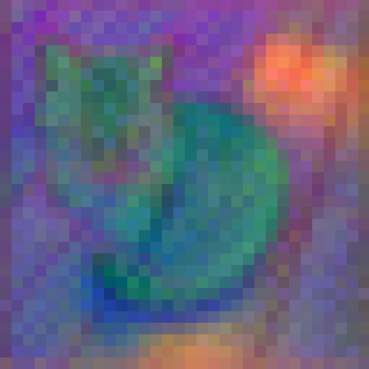} \\

  \end{tabular}


\vspace{6pt}
\makebox[\textwidth][c]{
  \hspace*{6.4cm}\parbox{0.7\textwidth}{
  }
}

\end{minipage}
\hfill
\begin{minipage}[t]{0.49\textwidth}
  \vspace{0pt}%
  \centering
  \footnotesize
\begin{tikzpicture}[baseline=(current bounding box.north)]
    \hspace{5pt}
    \begin{axis}[
      scale only axis,
      width=0.8\linewidth,
      height=0.55\linewidth,
      xmin=3.9, xmax=10.3,
      ymin=1.8,  ymax=5.7,
      xtick={3.9120,4.6052,5.2983,5.9915,7.3659,8.6153,9.4453,10.2848},
      xticklabels={,100K,,400K,1M,2.3M,4M,7M},
      ytick={1.8326,3.2189,4.3175,5.0106,5.7038},
      yticklabels={5,10,20,40,80},
      xlabel={\textbf{Training Iteration}},
      ylabel={\textbf{\Th{gFID}-50K}},
        y label style={
            at={(axis description cs:-0.09,0.6)}, 
            anchor=east                         
        },
        legend style={
            draw=black,                 
            line width=0.5pt,             
            at={(0.425,0.996)},
            anchor=north west
        },
      grid=major,  
      major grid style={gray!20},
    ]

      \addplot[color=teal!30, mark=*, line width=1.2pt] 
        coordinates {
           (3.9120, 5.7038)
           (4.6052, 5.2136)
           (5.2983, 4.7230)
           (5.9915, 4.2922)
           (8.6153, 3.3542)
           (10.2848, 3.1372)
        };
      \addlegendentry{\texttt{DiT-XL/2}}

      \addplot[color=teal, mark=*, line width=1.2pt]
        coordinates {
           (3.9120, 5.6326)
           (4.6052, 5.0081)
           (5.2983, 4.4485)
           (5.9915, 3.9620)
           (7.3659, 3.1988)
        };
      \addlegendentry{\texttt{DiT-XL/2 + Ours}}

      \addplot[color=softblue!30, mark=*, line width=1.2pt]
        coordinates {
           (3.9120, 5.2788)
           (4.6052, 4.2870)
           (5.2983, 3.4276)
           (5.9915, 2.7474)
           (7.3659, 2.3263)
           (9.4453, 2.1636)
        };
      \addlegendentry{\texttt{REPA}}

      \addplot[color=softblue, mark=*, line width=1.2pt]
        coordinates {
           (3.9120, 5.2054)
           (4.6052, 4.2395)
           (5.2983, 3.3354)
           (5.9915, 2.6435)
           (7.3659, 2.1534)
        };
      \addlegendentry{\texttt{REPA + Ours}}

      \draw[<-, dashed, teal]
        (axis cs:7.45, 3.1988) -- (axis cs:10.2848, 3.1372)
        node[midway, below=-0.7mm] {\(\times 7\) speed-up};

      \draw[<-, dashed, softblue]
        (axis cs:7.45, 2.1534) -- (axis cs:9.4453, 2.1636)
        node[midway, below=-0.7mm] {\(\times 4\) speed-up};

    \end{axis}
\end{tikzpicture}


\vspace{0.5pt}
\makebox[\textwidth][c]{
  \hspace*{6.4cm}\parbox{0.7\textwidth}{
  }
}

\end{minipage}

\vspace{-5pt}

\caption{
\textbf{Latent Space Structure (Left)} Top three principal components of \sdvae and \texttt{SDXL-VAE}, with and without \our, demonstrating visually that our regularization produces smoother latent representations without compromising reconstruction (See ~\autoref{tab:comp_auto}). 
\textbf{Accelerated Training (Right)} Training curves (without classifier-free guidance) for \ditxltwo and \texttt{REPA (w/ SiT-XL/2)}, 
showing that our \our accelerates convergence by $\times 7$ and $\times 4$, respectively. 
}
\label{fig:teaser}
\end{figure*}
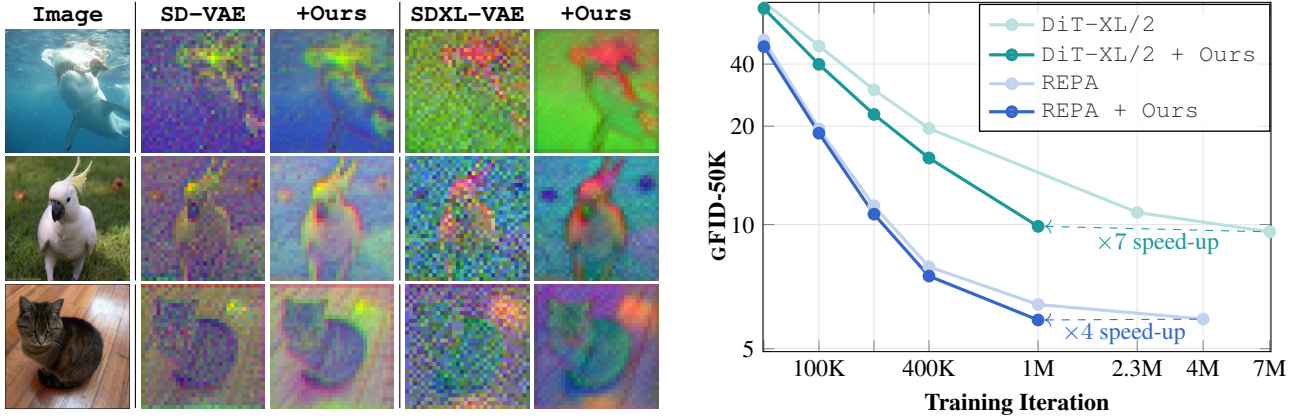

Latent generative models \cite{rombach2022high} have become a dominant framework for high-fidelity image synthesis, achieving state-of-the-art results across diffusion models \cite{rombach2022high, yao2024fasterdit, ma2024sit}, masked generative modeling \cite{chang2022maskgit, li2023mage}, and autoregressive models \cite{esser2021taming, li2024autoregressive, tian2024visual}. These models operate in two phases. First, an autoencoder compresses high-dimensional images into a lower-dimensional latent space, which can be continuous (e.g., \texttt{SD-VAE} for diffusion \cite{rombach2022high}) or discrete (e.g., \texttt{VQ-GAN} for autoregressive \cite{esser2021taming, yu2022scaling} and masked generative modeling \cite{chang2022maskgit}). This latent space retains essential semantic and structural information while discarding high-frequency details. Second, a generative model learns to model the distribution of these latent representations, enabling the synthesis of visually coherent images. At inference time, the generative model first samples a latent code, which is then decoded back into the image space by the autoencoder. While much research has focused on improving the generative phase—through advances in architectures \cite{peebles2023scalable}, objectives \cite{ma2024sit}, and optimization techniques \cite{yao2024fasterdit}—the autoencoder's role in shaping the latent space remains equally critical to overall performance.

In fact, the quality of the latent space is pivotal, influencing both computational efficiency (by reducing dimensionality and accelerating convergence in the generative phase) and the model's ability to produce high-fidelity outputs \cite{rombach2022high}. 
In diffusion models, most state-of-the-art approaches—such as \texttt{DiT} \cite{peebles2023scalable}, \texttt{SiT} \cite{ma2024sit}, \texttt{PixArt} \cite{chen2024pixartalpha}, \texttt{SD3} \cite{esser2024sd3}, and \texttt{Flux} \cite{flux2023}—rely on autoencoders with architectures and training objectives similar to the \texttt{SD-VAE} introduced in Latent Diffusion Models (LDM) \cite{rombach2022high}. LDM explores two widely adopted regularization strategies: a continuous variational approach and a discrete codebook framework. The variational approach uses a KL divergence term to align the latent distribution with a Gaussian prior, promoting a smooth and structured latent space \cite{kingma2014}. Alternatively, the discrete codebook framework constrains the latent space to a finite set of learned embeddings, limiting its complexity and providing a different form of regularization \cite{esser2021taming}.

These regularization strategies inherently introduce a trade-off. Stronger regularization, such as increasing the weight of the KL divergence term, produces a smoother and more learnable latent space for the generative model in the second phase \cite{tschannen2025givt}. 
However, it also reduces the information capacity of the latent representation, leading to a loss of fine-grained details and ultimately degrading reconstruction quality. Empirical evidence suggests that this trade-off can set an upper bound on the overall performance of latent generative models \cite{rombach2022high}, as the autoencoder’s limited capacity to preserve detailed information restricts the overall ability of latent generative models to synthesize highly-fidelity images. This raises a fundamental question:  
\emph{Can we mitigate this trade-off, creating a latent space that is more optimized for generative modeling, without compromising reconstruction quality, thereby improving the overall generative modeling process?}

A key aspect that could address this challenge lies in the structure and properties of the latent space itself. In particular, we identify an essential limitation of current state-of-the-art autoencoders: their latent representations are not equivariant to basic spatial transformations, such as scaling and rotation (see \autoref{fig:qualitative-equivariance-main}; extended discussion in \secref{sec:irregularity}).
This introduces unnecessary complexity into the latent manifold, forcing the generative model to learn nonlinear relationships that could otherwise be avoided.

To address this issue, we propose a simple yet effective modification to the training objective of autoencoders that encourages latent spaces to exhibit the aforementioned equivariance.
Our method called \texttt{EQ-VAE}, penalizes discrepancies between reconstructions of transformed latent representations and the corresponding transformations of input images. Notably, \texttt{EQ-VAE} requires no architectural changes to existing autoencoder models and does not necessitate training from scratch. 
Instead, fine-tuning pre-trained autoencoders for a few epochs with \texttt{EQ-VAE} suffices to imbue the latent space with equivariance properties, reducing its complexity (see \autoref{fig:teaser}-left; quantitative results in \autoref{tab:ablation-trans}) and facilitating learning for generative models (e.g., \autoref{fig:teaser}-right). 
This is achieved without degrading the autoencoder’s reconstruction quality.

Our method is compatible with both continuous and discrete autoencoders, enabling broad applicability across latent generative models. For example, applying \texttt{EQ-VAE} to the continuous \texttt{SD-VAE} \cite{rombach2022high} significantly improves the performance of downstream diffusion models such as \texttt{DiT} \cite{peebles2023scalable}, \texttt{SiT} \cite{ma2024sit}, and \texttt{REPA} \cite{Yu2025repa}, as measured by FID scores. 
Similarly, applying \texttt{EQ-VAE} to discrete \texttt{VQ-GAN} \cite{esser2021taming} enhances performance in the masked generative modeling approach \texttt{MaskGIT} \cite{chang2022maskgit}.

We make the following contributions:
\begin{itemize}[noitemsep,topsep=0pt]
    \item We identify that the latent space of established autoencoders lacks equivariance under spatial transformations, which impedes latent generative modeling. Building on this observation, we propose \our, a simple regularization strategy that improves generative performance without compromising reconstruction quality.  
    \item Our method is compatible with both continuous and discrete autoencoders, enabling a plug-and-play approach for commonly used generative models such as diffusion and masked generative models.
    \item 
    We show that by fine-tuning well-established autoencoders with our objective,
    we significantly accelerate the training of latent generative models. For instance, fine-tuning \sdvae for just $5$ epochs yields a $\times 7$ speedup on \ditxltwo and  $\times 4$ speedup on \texttt{REPA (w/ SiT-XL/2)} (see \autoref{fig:teaser} (right)).
\end{itemize}
\section{Related work}
\label{sec:related}

\paragraph{Autoencoders for Latent Generative Models}
Training diffusion models directly in pixel space is computationally inefficient, as most of the bits in a digital image correspond to subtle details with little perceptual significance. To overcome this issue, \citet{rombach2022high} propose latent diffusion models that operate in a compressed latent space produced in a separate stage by an autoencoder. Their KL-regularized autoencoder, \texttt{SD-VAE}, has been extensively utilized in numerous diffusion models \cite{yao2024fasterdit, ma2024sit, chen2024pixartalpha}. Subsequent research has primarily focused on minimizing the reconstruction error that sets an upper bound on generative performance,
by increasing the number of latent channels \cite{esser2024sd3, flux2023, dai2023emu} and incorporating task specific priors \cite{zhu2023designing}.
To enable efficient training on high-resolution images \citet{xie2025sana} and \citet{chen2025deep}  extensively increase the compression ratio without compromising the reconstruction quality.
\citet{hu2023complexity} investigate the ideal latent space for generative models and find that a relatively weak decoder produces a latent distribution that enhances generative performance. 
Discrete autoencoders are initially introduced with \texttt{VQ-VAE} \cite{oord2017vq} to quantize image patches into discrete visual tokens. \texttt{VQ-GAN} \cite{esser2021taming} further refines \texttt{VQ-VAE} by integrating adversarial and perceptual losses,
enabling more accurate and detailed representations. Subsequent works have focused on architectural improvements \cite{yu2022vectorquantized}, strategies to increase the codebook size and maximize its utilization \cite{yu2024language, zhu2024scaling}.
Unlike these prior approaches, we investigate a novel perspective—leveraging spatial equivariance—to shape a latent space better suited for generative modeling.

\paragraph{Auxiliary Objectives and Regularization in VAEs}
 Autoencoders are designed to learn latent spaces that compactly represent meaningful features of the observed data. However, without any regularization, their latent code lacks meaningful structure. Variational Autoencoders (VAEs) were introduced in \citet{kingma2014} to address this by minimizing the KL divergence between the latent distribution and a Gaussian prior. Many subsequent works have adopted and extended this framework \cite{Higgins2016betaVAELB, dilokthanakul2016deep, tomczak2018vae, takahashi2019variational}.
Other works have proposed alternative regularizations based on the Wasserstein distance \cite{tolstikhin2018wasserstein, kolouri2018sliced}, adversarial objectives \cite{zhao2018adversa, makhzani2015adversarial} and vector quantization (VQ) \cite{oord2017vq}. Closely related to our work, \citet{NEURIPS2021_6c19e0a6} proposes a consistency regularization enforcing the latent code to be invariant under spatial transformations.  Our \texttt{EQ-VAE} promotes \emph{equivariance} rather than invariance under spatial transformations and we extensively demonstrate the impact of equivariance regularization on latent generative modeling.

\paragraph{Equivariance in Computer Vision} The success of Convolutional neural networks (CNN) in numerous computer vision tasks can be largely attributed to their approximate translation equivariance that arises due to the nature of convolution.
To incorporate other symmetries in the data, various group-equivariant convolutional networks have been proposed, including roto-translation equivariance in 2D \cite{cohen2016group, marcos2017rotation, hoogeboom2018hexaconv, weiler2019general}, extensions in 3D \cite{worrall2018cubenet, thomas2018tensor, kondor2018n}, and scale equivariance \cite{rahman2023truly, Sosnovik2020Scale-Equivariant}. The derivation of group equivariance constraint typically results in steerable filters constructed from a basis. Besides architectural constraints, equivariance can be achieved by parameter sharing \cite{ravanbakhsh2017equivariance}, frame averaging \cite{puny2022frame}, and canonicalization functions \cite{kaba2023equivariance}.  For autoencoder models, \citet{winter2022unsupervised} produce latent representations
that are separated into a group invariant and equivariant part. Closely related to our work \citet{ryu2024vqgan}, train autoencoders to be equivariant under horizontal and vertical flips.
However, they do not investigate the impact of equivariant representations on latent generative modeling.

\section{Method}
\label{sec:equi-vae}

This section presents our methodology. We first provide an overview of autoencoder models for latent generative modeling (\secref{sec:prelim}), focusing on the continuous case used in diffusion models. We then highlight the lack of equivariance in latent representations (\secref{sec:irregularity}) and introduce EQ-VAE to address it (\secref{sec:method-eqvae}).

\subsection{Preliminary: Continuous Autoencoders for Latent Generative Modeling}
\label{sec:prelim}

The first modeling stage consists of an autoencoder that compresses the pixel space into a continuous (\citet{rombach2022high}) or discrete (\citet{esser2021taming}) latent space. We focus here on the continuous case.
Given an input image $\mathbf{x} \in \mathbb{R}^{H \times W \times 3}$, an 
encoder $\mathcal{E}$ transforms the image into a compressed representation $\mathbf{z}= \mathcal{E(\mathbf{x})} \in \mathbb{R}^{\frac{H}{f} \times \frac{W}{f} \times c}$, where $f$ is the compression ratio and $c$ are the latent channels. Then a decoder $\mathcal{D}$ takes as input the latent representation and reconstructs the image $\hat{\mathbf{x}} = \mathcal{D}(\mathbf{z})$. For an input image $\mathbf{x}$ the training objective reads as follows:
\begin{align}
    \label{eq:ldm}
    \mathcal{L}_{\text{VAE}}(\mathbf{x}) = 
    \mathcal{L}_{rec}(\mathbf{x}, \mathbf{\hat{x}}) + \lambda_{gan} \mathcal{L}_{gan}(\mathbf{\hat{x}}) + 
    \lambda_{reg} \mathcal{L}_{reg}
\end{align}
where $\mathcal{L}_{rec}$ consists of a pixel space reconstruction objective and a perceptual loss such LPIPS \cite{zhang2018unreasonable}, $\mathcal{L}_{gan}$ is a patch-based adversarial loss \cite{isola2017image}
and $\mathcal{L}_{reg}$ is usually a Kullback-Leibler regularization with a Gaussian prior \cite{kingma2014}.

\begin{figure}[t]
    \centering
    \setlength{\tabcolsep}{1.5pt}
    \begin{tabular}{cccc}
      \multirow{2}{*}{\small Input Image \textbf{x}}
      & \multicolumn{2}{c}{{\small \texttt{SD-VAE}}}
      & {\texttt{Ours}} \\
      \cmidrule(lr){2-3} \cmidrule(lr){4-4}
       
      & {\small$\mathcal{D} ( \mathcal{E}(\tau \circ \mathbf{x}))$}
      & {\small$\mathcal{D}(\tau \circ \mathcal{E}(\mathbf{x}))$}
      & {\small$\mathcal{D}(\tau \circ \mathcal{E}(\mathbf{x}))$} \\
      
      \vspace{-0.4cm} \\
      
     \includegraphics[width=0.225\linewidth]{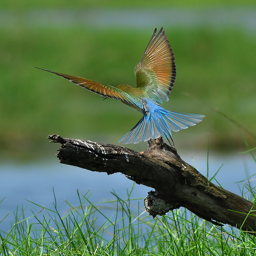} 
      & \includegraphics[width=0.225\linewidth]{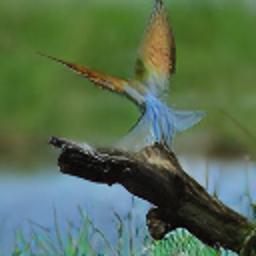} 
      & \includegraphics[width=0.225\linewidth]{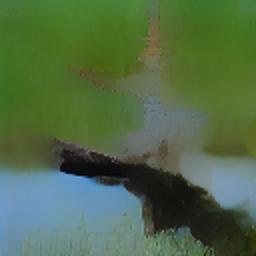} 
      & \includegraphics[width=0.225\linewidth]{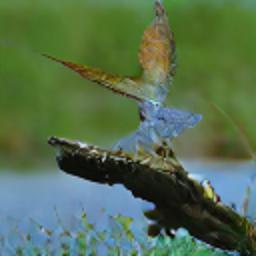} \\
      
      \includegraphics[width=0.225\linewidth]{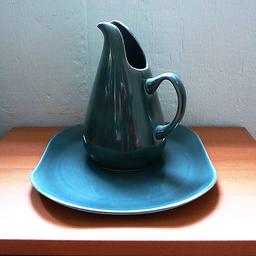} 
      & \includegraphics[width=0.225\linewidth]{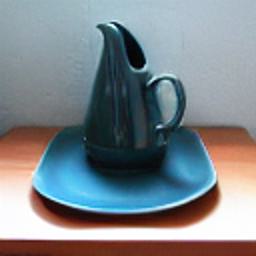} 
      & \includegraphics[width=0.225\linewidth]{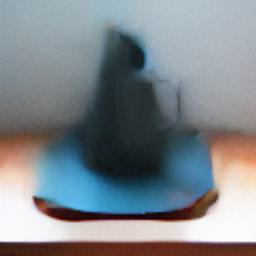} 
      & \includegraphics[width=0.225\linewidth]{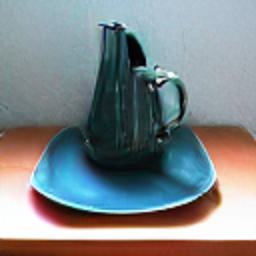} \\

    \end{tabular}
  \caption{
  \textbf{Latent Space Equivariance.} 
  Reconstructed images using \texttt{SD-VAE}~\cite{rombach2022high} and our \our when applying scaling transformation $\tau$, with factor $s=0.5$, to the input images $\mathcal{D}(\mathcal{E}(\tau \circ \mathbf{x}))$ versus directly to the latent representations $\mathcal{D}(\tau \circ \mathcal{E}(\mathbf{x}))$. Our approach preserves reconstruction quality under latent transformations, whereas \texttt{SD-VAE} exhibits significant degradation. See ~\autoref{fig:qualitative-equivariance-appendix} for additional examples.
  }
  \label{fig:qualitative-equivariance-main}
\end{figure}


\definecolor{SDVAE}{RGB}{0,128,128}          
\definecolor{SDVAEOurs}{RGB}{0,100,100  }     
\definecolor{SDXLVAE}{RGB}{44, 62, 80}        
\definecolor{SDXLVAEOurs}{RGB}{255, 140, 0}   

\begin{figure}[ht]
    \centering
    \begin{tikzpicture}
        \begin{axis}[
            ybar,
            width=8.5cm,
            height=5cm,
            bar width=9pt,
            ymin=0,
            ymax=220,
            ylabel={\Th{\textbf{rFID}}},
            y label style={
                at={(axis description cs:-0.12,0.6)}, 
                anchor=east                         
            },
            xlabel={\textbf{Transformations}},
            symbolic x coords={$s=0.25$,$s=0.50$,$s=0.75$,$\theta$},
            xtick=data,
            ymajorgrids=true,
            grid style=dashed,
            enlarge x limits=0.15,
            legend style={
                at={(0.5,1.05)},
                anchor=south,
                legend columns=2,
            },
            nodes near coords,
            every node near coord/.append style={font=\footnotesize, anchor=south},
            label style={font=\small},
            ticklabel style={font=\small},
        ]
        
        \addplot [
            fill=SDVAE!50, 
            draw=black, 
        ] coordinates {
            ($s=0.75$, 52)
            ($s=0.50$, 83)
            ($s=0.25$, 107)
            ($\theta$, 78)
        };
        
        \addplot [
            fill=SDVAE, 
            draw=black, 
        ] coordinates {
            ($s=0.75$, 2)
            ($s=0.50$, 7)
            ($s=0.25$, 31)
            ($\theta$, 12)
        };
        
        \addplot [
            fill=SDXLVAE, 
            draw=black, 
        ] coordinates {
            ($s=0.75$, 64)
            ($s=0.50$, 117)
            ($s=0.25$, 130)
            ($\theta$, 192)
        };
        
        \addplot [
            fill=SDXLVAE!50, 
            draw=black, 
        ] coordinates {
            ($s=0.75$, 3)
            ($s=0.50$, 10)
            ($s=0.25$, 34)
            ($\theta$, 12)
        };
        
        \legend{SD-VAE, SD-VAE+(ours), SDXL-VAE, SDXL-VAE+(ours)}
        
        \end{axis}
    \end{tikzpicture}
    \caption{\textbf{Enhanced Reconstruction under Latent Transformations.} Reconstruction \Th{rFID} measured between $\tau \circ \mathbf{x}$ and $ \mathcal{D}(\tau \circ \mathcal{E}(\mathbf{x}))$ for various spatial transformations. We consider scaling transforms with factors $s = 0.75, 0.50, 0.25$ and also measure the average \Th{rFID}  over rotation angles $\theta = \frac{\pi}{2}, \pi, \frac{3\pi}{2}$. Results for \sdvae~\cite{rombach2022high} and \texttt{SDXL-VAE}~\cite{podell2024sdxl}, with and without \our. Our approach significantly reduces \Th{rFID} compared to baselines, improving image fidelity under latent transformations. For readability, we show $\floor{\Th{rFID}}$. }
    \label{fig:scales-rfig}
\end{figure}
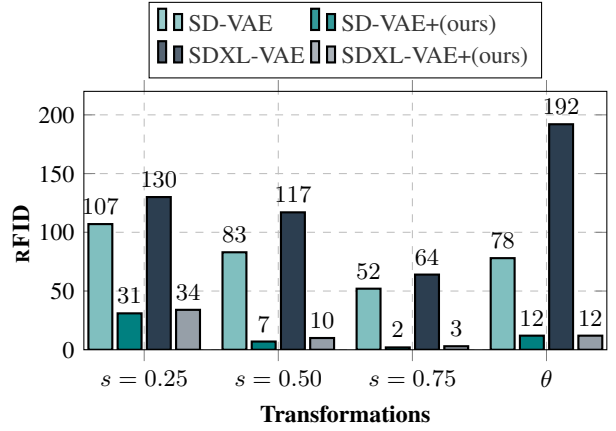

\subsection{Lack of Equivarance under Spatial Tansformations}
\label{sec:irregularity}

Our work is motivated by a key observation: state-of-the-art autoencoders, such as \texttt{SD-VAE} \cite{rombach2022high}, produce latent representations 
$\mathcal{E}(\mathbf{x})$ that are not equivariant under basic spatial transformations like scaling and rotation. 

We formalize this as follows:

\noindent \textbf{Spatial Transformation} Let $\mathbf{x}(\mathbf{p}): \mathbb{R}^2 \rightarrow \mathbb{R}^c$ be an image (or latent representation) defined over 2D coordinates  \mbox{$\mathbf{p} = [u, v]^\top$}. 
A spatial transformation $\mathbf{\tau} \in \mathbb{R}^{2 \times 2}$ acts on the coordinates $p$ transforming $\mathbf{x}$ as follows:
\begin{equation} \label{eq:transformation}
\mathbf{x}_{\tau}(\mathbf{p}) =\mathbf{x}(\mathbf{\tau}^{-1} \mathbf{p})\text{,}
\end{equation}
denoted compactly for all $\mathbf{p}$ as $\tau \circ \mathbf{x}$.

\noindent \textbf{Equivariance} A latent representation $\mathcal{E}(\mathbf{x})$ is equivariant with a
transformation $\tau$ of the input image $\mathbf{x}$ if the transformation
can be transferred to the representation output: 
\begin{align}
    \label{eq:equivariace}
    \forall \mathbf{x} \in \mathcal{X}: \quad \mathcal{E}(\tau \circ \mathbf{x}) = \tau \circ \mathcal{E}(\mathbf{x})\text{.}
\end{align}

To test whether the latent representations of autoencoder models are equivariant under spatial transformations, we applied scaling and rotations $\tau$ directly to the latent code and evaluated the corresponding reconstructions. 
Specifically, we compare decoding transformed latent representations, $\mathcal{D}(\mathbf{\tau} \circ \mathcal{E}(\mathbf{x}))$, to decoding latents of transformed input images, $\mathcal{D}(\mathcal{E}(\mathbf{\tau} \circ \mathbf{x}))$. 
We present qualitative and quantitative results in \autoref{fig:qualitative-equivariance-main} and \autoref{fig:scales-rfig} respectively.

Our findings reveal a clear disparity: while autoencoders reconstruct images accurately when transformations are applied to the input (i.e., $\mathcal{D}(\mathcal{E}(\mathbf{\tau} \circ \mathbf{x}))$), applying transformations directly to the latent representation (i.e., $\mathcal{D}(\mathbf{\tau} \circ \mathcal{E}(\mathbf{x}))$) leads to significant degradation in reconstruction quality.

This limitation arises because (1) convolutional architectures commonly used in the autoencoders of latent generative models, such as \texttt{SD-VAE}, are not equivariant under arbitrary spatial transformations such as scaling and rotation, and (2) their standard training objectives (for example, reconstruction loss and KL divergence) do not explicitly or implicitly encourage equivariance. As a result, semantically similar inputs, such as an image $\mathbf{x}$ and its scaled counterpart $\mathbf{\tau} \circ \mathbf{x}$, are encoded into latent codes $\mathcal{E}(\mathbf{x})$ and $\mathcal{E}(\mathbf{\tau} \circ \mathbf{x})$ that are not related by the corresponding spatial transformation, i.e. $\mathcal{E}(\mathbf{\tau} \circ \mathbf{x}) \neq \mathbf{\tau} \circ \mathcal{E}(\mathbf{x})$, thus unnecessarily complicating the structure of the latent space.

\subsection{EQ-VAE: Regularization via equivariance constraints}
\label{sec:method-eqvae}

To address this limitation, we propose \texttt{EQ-VAE}, which regularizes the latent representations to promote equivariance under spatial transformations.
As seen in \autoref{fig:teaser} (left) this produces smoother latent representations, enabling more efficient learning.

\noindent \textbf{Explicit Regularization.} 
A direct way to enforce equivariance is to include the equivariance constraint from \Equation{eq:equivariace} as a loss term during training:
\begin{equation}
    \label{eq:encoder_loss}
     \mathcal{L}_{\text{explicit}}(\mathbf{x}) = 
     \Vert \tau \circ \mathcal{E}(\mathbf{x}) - \mathcal{E}(\tau \circ  \mathbf{x}) \Vert_2^2 
     \text{,}
\end{equation}
where $\tau$ is sampled from a set of spatial transformations. However, minimizing this loss alone can lead to trivial solutions, such as collapsing the latent representation to a constant value $\mathcal{E}(\mathbf{x}) = \text{c}, \; \forall \mathbf{x}$, which we observe in our experiments (see \autoref{tab:losses}), making explicit regularization ineffective.

\noindent \textbf{Implicit Regularization.} 
To overcome this limitation of explicit regularization, we adopt an implicit approach. Inspired by the findings in \autoref{fig:qualitative-equivariance-main}, this approach aligns the reconstructions of transformed latent representations ($\mathcal{D}\big( \tau \circ \mathcal{E}(\mathbf{x}) \big)$) with the corresponding transformed inputs ($\tau \circ \mathbf{x}$ ). Specifically, we modify the original training objective of \Equation{eq:ldm} as follows:
\begin{align}
    \label{eq:ours_obj}
    \mathcal{L}_{\text{EQ-VAE}} (\mathbf{x}, {\color{DarkenedMagenta}{\tau}}) = 
     \mathcal{L}_{rec}& \Big({\color{DarkenedMagenta}{\text{$\mathbf{\tau} \circ$}}} \mathbf{x}, \mathcal{D}\big ({\color{DarkenedMagenta}{\text{$\mathbf{\tau} \circ$}}}\mathcal{E}(\mathbf{x}) \big) \Big) + \\
    \lambda_{gan} \mathcal{L}_{gan}& \Big( \mathcal{D}\big ({\color{DarkenedMagenta}{\text{$\mathbf{\tau} \circ$}}} \mathcal{E}(\mathbf{x}) \big) \Big) + \lambda_{reg} \mathcal{L}_{reg} \notag
\end{align}
where the changes compared to \Eq{eq:ldm} are highlighted in {\color{DarkenedMagenta}{color}}.
Notice that when $\tau$ is the identity transformation, this formulation reduces to the original objective in \Eq{eq:ldm}. By leveraging the rich supervision signal from both reconstruction and adversarial objectives, this approach implicitly encourages the encoder to produce equivariant latent representations while avoiding mode collapse (see \secref{sec:appenidx_exp_imp}).

\noindent \textbf{Transformation Design.} 
We focus on two types of spatial transformations: anisotropic scaling and rotations. These are parameterized as:
\begin{equation}
\mathbf{S}(s_x, s_y) =
\begin{bmatrix}
s_x  & 0 \\
0 &  s_y
\end{bmatrix}    
,\quad 
\mathbf{R}(\theta)=
\begin{bmatrix}
 \cos\theta & -\sin\theta \\
\sin\theta &   \cos\theta
\end{bmatrix}    
\end{equation}
The final transformation is the composition of scaling and rotation: $\tau = \mathbf{S}(s_x, s_y) \cdot \mathbf{R}(\theta)$.
We sample uniformly $0.25 < s_x, s_y < 1$, and $\theta \in ( \frac{\pi}{2},\pi, \frac{3\pi}{2})$. We consider these three rotation angles (multiples of 90$^{\circ}$) to avoid corner artifacts. For downsampling, we use bicubic interpolation. Empirically, we find \emph{scaling equivariance is more beneficial} for generation than rotation equivariance (see \autoref{tab:ablation-trans}).

To preserve the prior reconstruction capabilities of the autoencoder, we return to the standard objective (\Eq{eq:ldm}) by sampling the identity transform \(\tau = \mathbf{I}\) in \Eq{eq:ours_obj} with probability \(p_{\alpha}\). 
Our total objective can thus be written as:
\
\begin{align}
\mathcal{L}_{total}(\mathbf{x}) = 
\begin{cases}
    \mathcal{L}_{\text{VAE}}(\mathbf{x}) \qquad  & p <p_{\alpha}, \\
   \mathcal{L}_{\text{EQ-VAE}}(\mathbf{x}, \tau) \qquad &   p \geq p_{\alpha} . \\
\end{cases}
\end{align}
where $p$ is  sampled uniformly from $[0, 1]$. This controls the strength of our regularization. By default we set $p_{\alpha}=0.5$  (we ablate regularization strength in \secref{sec:appendix_prior}).

We note that our approach enforces equivariance by applying transformations directly to the latent space, distinguishing it from methods relying on input data augmentation \cite{brehmer2024does}.

\noindent \textbf{Extending EQ-VAE to Discrete Autoencoders.} So far, we described \our in the context of continuous autoencoders.
In discrete autoencoders e.g., \texttt{VQ-GAN}~\cite{esser2021taming}, the encoder outputs continuous features $\mathcal{E}(\mathbf{x})$ that are mapped to the nearest entry in a learned codebook, forming a discretized latent space via quantization. 
Adapting our method for discrete autoencoders, such as \texttt{VQ-GAN}, is straightforward. 
We employ our equivariance regularization loss as described in \secref{sec:method-eqvae} and apply the transformations $\tau$ on the latent features $\mathcal{E}(\mathbf{x})$ \emph{before} the quantization.

\section{Experiments}
\label{sec:experiments}

\subsection{Setup}
\paragraph{Implementation Details}
We finetune all autoencoders on OpenImages to adhere to the framework used in LDM \cite{rombach2022high}. We finetune for $5$ epochs with batch size $10$.  Detailed specifications of each autoencoder, including spatial compression rates and latent channels, are provided in \autoref{sec:appendix:ae_specs}. 
For \texttt{DiT}~\cite{peebles2023scalable}, \texttt{SiT}~\cite{ma2024sit} and \texttt{REPA}~\cite{Yu2025repa}, we follow their default settings and train on ImageNet~\cite{deng2009imagenet} with a batch size of $256$, where each image is resized to $256\times 256$.
We use $\text{B/}2$, $\text{XL/}2$  architectures which employ a patch size $2$, except for the experiment with \texttt{SD-VAE-16} in~\autoref{tab:comp_auto} in which we used $\text{B/}1$, due to its lower spatial resolution compared to other autoencoders.
These models are originally trained in the latent distribution of \texttt{SD-VAE-FT-EMA}\footnote{https://huggingface.co/stabilityai/sd-vae-ft-ema} a subsequent version of the original \texttt{SD-VAE} that has been further fine-tuned with an exponential moving average on LAION-Aesthetics~\cite{schuhmann2022laion} (see \autoref{tab:ablation_ft} and \cite{peebles2023scalable} for their performance differences). 
For \texttt{MaskGIT}, we follow \cite{besnier2023pytorch} and train on ImageNet for 300 epochs with a batch size of $256$. 
We follow \texttt{ADM} \cite{dhariwal2021adm} for all data pre-processing protocols.

\begin{table}[t]
\footnotesize
\centering
\setlength{\tabcolsep}{3pt} 
\begin{tabular}{llcccc}
\toprule
& \multirow{2}{*}{\Th{Autoencoder}}  & \multirow{2}{*}{\Th{rFID$\downarrow$}} & \multirow{2}{*}{\Th{gFID$\downarrow$}}  & \mc{2}{\Th{Equiv. Error}}  \\  \cmidrule(lr){5-6}
& & &  & $R(\theta)$\Th{$\downarrow$} & $S(s)$\Th{$\downarrow$} \\ 
\midrule 
 & \sdvae  & $0.90$ & $43.8$ & $0.89$ & $0.69$ \\
 & \cellcolor{TableColor} + \our (ours)   &\cellcolor{TableColor}$0.82$ &\cellcolor{TableColor}$34.1$ &\cellcolor{TableColor}$0.56$ & \cellcolor{TableColor}$0.43$ \\ 
\graycline{2-6}
\vspace{-10.0pt}
\multirow{6}{*}{%
  \makebox[0pt][c]{\rotatebox{90}{\Th{Cont.}}}%
}
 & & & & &\\

 & \sdxlvae  & $0.67$ & $46.0$ & $1.25$ & $0.97$  \\
 & \cellcolor{TableColor} + \our (ours) &\cellcolor{TableColor}$0.65$  & \cellcolor{TableColor}$35.9$ &\cellcolor{TableColor}$0.65$ &\cellcolor{TableColor}$0.35$  \\ 
\graycline{2-6}

 & \texttt{SD3-VAE}  &  $0.20$ & $58.9$  & $0.51$ & $0.16$  \\
 & \cellcolor{TableColor} + \our (ours)  & \cellcolor{TableColor}$0.19$ & \cellcolor{TableColor}$54.0$ & \cellcolor{TableColor}$0.37$ & \cellcolor{TableColor}$0.11$  \\ 

\graycline{2-6}

 & \texttt{SD-VAE-16}  &  $0.87$ & $64.1$ & $0.95$ & $0.85$  \\
 & \cellcolor{TableColor} + \our (ours)  & \cellcolor{TableColor}$0.82$ &\cellcolor{TableColor}$49.7$ &\cellcolor{TableColor}$0.39$ &\cellcolor{TableColor}$0.17$  \\ 

\hline\hline
 

\multirow{2}{*}{\rotatebox{90}{\Th{Disc.}}} & \texttt{VQ-GAN} & $7.94$  & ~~$6.8$ & $1.35$ & $1.22$ \\
 & \cellcolor{TableColor} + \our (ours)  & \cellcolor{TableColor}$7.54$  &\cellcolor{TableColor}~~$5.9$ & \cellcolor{TableColor}$0.64$ & \cellcolor{TableColor}$0.55$ \\ 
\bottomrule
\end{tabular}
\vspace{-3pt}
\caption{\textbf{Comparison of Autoencoders with and without \our.} 
We evaluate reconstruction quality, equivariance errors (defined in \autoref{sec:appendix_metrics}), and generative performance for continuous (\texttt{SD-VAE}, \texttt{SDXL-VAE}, \texttt{SD3-VAE}) and discrete (\texttt{VQ-GAN}) autoencoders, with and without \our. 
Generative FID (\Th{gFID}) is 
measured using \texttt{DiT-B} for continuous VAEs and \texttt{MaskGIT} for \texttt{VQ-GAN}.
Our approach reduces reconstruction \Th{rFID} and equivariance errors while enhancing generative performance (\Th{gFID}). For additional reconstruction metrics see 
\autoref{tab:appendix_recon}.}
\label{tab:comp_auto}
\vspace{-0.5cm}
\end{table}
\vspace{-5 pt}
\paragraph{Evaluation}
For generative performance, we train latent generative models on the latent distribution of each autoencoder and we report Frechet Inception Distance (FID) \cite{fid}, sFID \cite{sfid}, Inception Score (IS) \cite{is}, Precision (Pre.) and Recall (Rec.) \cite{kynkaanniemi2019improved} using $50,000$ samples and following \texttt{ADM} evaluation protocol \cite{dhariwal2021adm}. 
To evaluate reconstruction, we report FID, Peak Signal-to-Noise Ratio (PSNR), Structural Similarity (SSIM) \cite{ssim}, and Perceptual Similarity (LPIPS) \cite{zhang2018unreasonable} using the ImageNet validation set. 
To distinguish reconstruction and generation FID, we write \Th{gFID} and \Th{rFID}, respectively. 
To quantify the effectiveness of
\our we further measure the equivariance error (see \autoref{sec:appendix_metrics}).

\begin{table}[t]
\footnotesize
\centering
\setlength{\tabcolsep}{4pt}
\begin{tabular}{lccc}
\toprule
\Th{Model} & \Th{\#Params} & \Th{Iter.} & \Th{gFID$\downarrow$} \\ 
\midrule 
\ditbtwo & $130\text{M}$ & $400\text{K}$ & $43.5$ \\
\rowcolor{TableColor} w/ \our (ours) & $130\text{M}$ & $400\text{K}$ & $34.1$ \\ 
\grayhline
\sitbtwo & $130\text{M}$ & $400\text{K}$ & $33.0$ \\
\rowcolor{TableColor} w/ \our (ours) & $130\text{M}$ & $400\text{K}$ & $31.2$ \\
\grayhline 
\ditxltwo & $675\text{M}$ & $400\text{K}$ & $19.5$ \\
\rowcolor{TableColor} w/ \our (ours) & $675\text{M}$ & $400\text{K}$ & $14.5$ \\ 
\grayhline
\sitxltwo & $675\text{M}$ & $400\text{K}$ & $17.2$ \\
\rowcolor{TableColor} w/ \our (ours) & $675\text{M}$ & $400\text{K}$ & $16.1$ \\
\midrule
\ditxltwo & $675\text{M}$ & ~~~$7\text{M}$ & ~~$9.6$ \\
\rowcolor{TableColor} w/ \our (ours) & $675\text{M}$ & $1.5\text{M}$ & ~~$8.8$ \\
\grayhline
\texttt{SiT-XL/2+REPA} & $675\text{M}$ & ~~~$4\text{M}$ & ~~$5.9$ \\
\rowcolor{TableColor} w/ \our (ours) & $675\text{M}$ & ~~~$1\text{M}$ & ~~$5.9$ \\ 
\bottomrule
\end{tabular}
\vspace{-3pt}
\caption{\textbf{\Th{gFID} Comparisons.} \Th{gFID} scores on ImageNet $256\times256$ for \texttt{DiT}, \texttt{SiT}, and \texttt{REPA} trained with either \texttt{SD-VAE-FT-EMA} or our \texttt{EQ-VAE}. No classifier-free guidance (CFG) is used. 
\texttt{EQ-VAE} consistently enhances both generative performance and training efficiency across all generative models.}
\label{tab:bench-diff-main}
\vspace{-3pt}
\end{table}

\subsection{Equivariance-regularized VAEs}
We begin our experimental analysis by demonstrating the versatility of \texttt{EQ-VAE}, showing that it seamlessly adapts to both continuous 
and discrete autoencoders.

\paragraph{Continuous Autoencoders}
We integrate our \texttt{EQ-VAE} regularization into established continuous autoencoders with varying latent dimensions. Namely,
\texttt{SD-VAE}, \texttt{SD-VAE-16}, \cite{rombach2022high}, \texttt{SDXL-VAE} \cite{podell2024sdxl}, and \texttt{SD3-VAE} \cite{esser2024sd3}. 
To evaluate the effect of the regularization on generative performance we train 
\texttt{DiT-B} models on the latent codes
before and after our regularization. 
We present our results in \autoref{tab:comp_auto}. 
We observe that our simple objective effectively reduces the equivariance error for all autoencoders. 
Further, \texttt{EQ-VAE} maintains the original autoencoders’ reconstruction fidelity while consistently delivering significant improvements in generative performance. 
The results hint that there is
a correlation between the generative performance (\Th{gFID}) and the reduction in equivariacne error. Notably, for \texttt{SD-VAE}, \texttt{SDXL-VAE} and \texttt{SD-VAE-16}, our regularization significantly boosts generative performance. For \texttt{SD3-VAE}, although the reduction in equivariance error is relatively modest, it still results in a \Th{gFID} improvement.

\begin{table}[t]
\footnotesize
\centering
\setlength{\tabcolsep}{3pt}
\begin{tabular}{lccc} 
\toprule
\Th{Model} & \Th{Epoch} & \Th{gFID$\downarrow$} & \Th{IS$\uparrow$}  \\ 
\midrule 
\texttt{MaskGIT}
  & $300$ 
  & $6.19$  
  & $182.1$ \\  

\grayhline
\texttt{MaskGIT}$^{\dagger}$
  & $300$ 
  & $6.80$  
  & $214.0$ \\  


\cellcolor{TableColor} w/ \our (ours)  
  &  \cellcolor{TableColor}$130$ 
  & \cellcolor{TableColor}$6.80$  
  & \cellcolor{TableColor}$188.1$ \\

\cellcolor{TableColor} w/ \our (ours) 
  & \cellcolor{TableColor}$300$ 
  & \cellcolor{TableColor}$5.91$  
  & \cellcolor{TableColor}$228.8$ \\ 

\bottomrule
\end{tabular}
\vspace{-3pt}
\caption{\textbf{Boosting Masked Generative Modeling.}  
Comparison of \Th{gFID} and \Th{IS} on ImageNet \(256\times256\) for \texttt{MaskGIT}~\cite{chang2022maskgit} and its open-source PyTorch reproduction$^{\dagger}$~\cite{besnier2023pytorch}, trained with either \texttt{VQ-GAN} or our \texttt{EQ-VAE}. \texttt{EQ-VAE} 
accelerates training by more than $\times 2$ (130 vs. 300 epochs), highlighting \texttt{EQ-VAE} can be effectively applied to vector-quantized autoencoders.}
\label{tab:maskgit}
\vspace{-3pt}
\end{table}

\paragraph{Discrete Autoencoders}
To investigate if \texttt{EQ-VAE} can be applied to discrete autoencoders, we experiment on \texttt{VQ-GAN} \cite{esser2021taming} and validate the effectiveness of our regularization on the masked image modeling framework \texttt{MaskGIT} \cite{chang2022maskgit}. In~\autoref{tab:comp_auto}, we show that \our is effective in the discrete case, reducing the equivariance error as well as improving the generative performance from $6.8$ to $5.9$ in \Th{gFID}.

\begin{figure*}[t]
    \centering
    \begin{tabular}{@{}l@{\hspace{2pt}}c@{\hspace{2pt}}c@{\hspace{2pt}}c@{\hspace{3pt}}c@{\hspace{2pt}}c@{\hspace{2pt}}c@{\hspace{3pt}}c@{\hspace{2pt}}c@{\hspace{2pt}}c@{}}
        & \text{50K} & \text{100K} & \text{400K} & \text{50K} & \text{100K} & \text{400K} & \text{50K} & \text{100K} & \text{400K} \\[0.35cm]

        \vspace{-1.6cm}
\rotatebox{90}{%
          \scriptsize
          \begin{tabular}{@{}l@{}}
            \multirow{2}{*}{\ditxltwo} \\
            
          \end{tabular}%
        } & & & & & & & & \\
         &
        \includegraphics[width=0.09\linewidth]{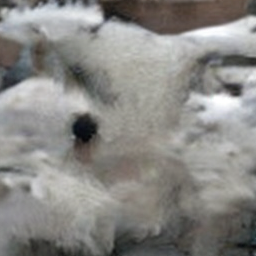} &
        \includegraphics[width=0.09\linewidth]{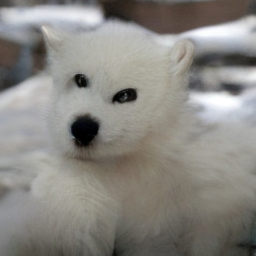} &
        \includegraphics[width=0.09\linewidth]{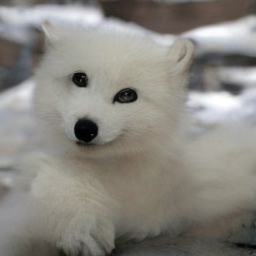} &
        \includegraphics[width=0.09\linewidth]{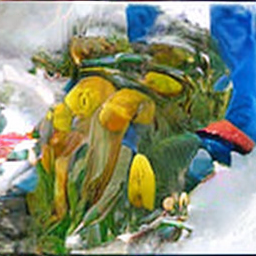} &
        \includegraphics[width=0.09\linewidth]{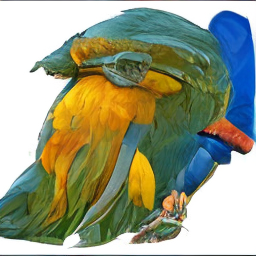} &
        \includegraphics[width=0.09\linewidth]{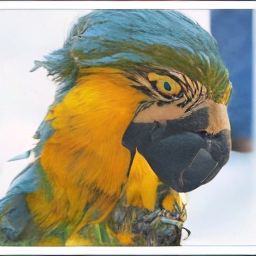} &
        \includegraphics[width=0.09\linewidth]{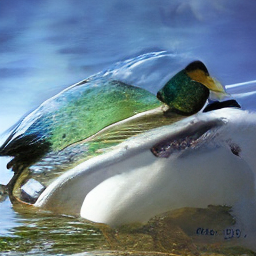} &
        \includegraphics[width=0.09\linewidth]{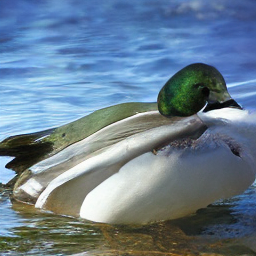} &
        \includegraphics[width=0.09\linewidth]{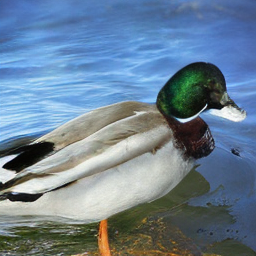} \\[0.2cm]
        
        \vspace{-1.6cm}
        \rotatebox{90}{%
          \scriptsize
          \begin{tabular}{@{}l@{}}
            \ditxltwo \\
            w/ \our 
          \end{tabular}%
        } & & & & & & & & \\ 
        
        &
        \includegraphics[width=0.09\linewidth]{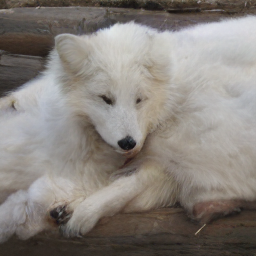} &
        \includegraphics[width=0.09\linewidth]{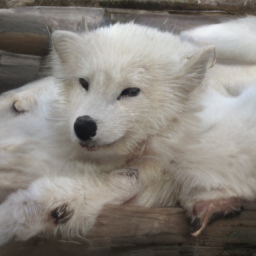} &
        \includegraphics[width=0.09\linewidth]{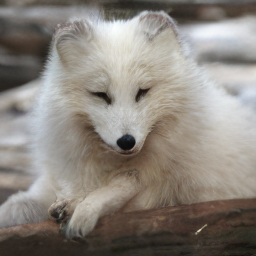} &
        \includegraphics[width=0.09\linewidth]{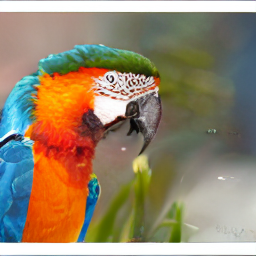} &
        \includegraphics[width=0.09\linewidth]{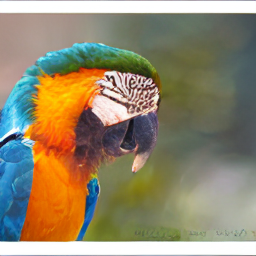} &
        \includegraphics[width=0.09\linewidth]{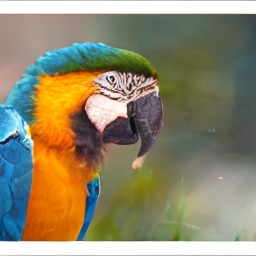} &
        \includegraphics[width=0.09\linewidth]{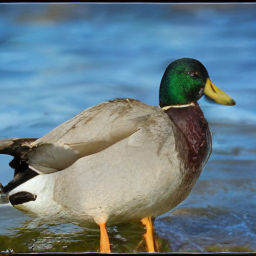} &
        \includegraphics[width=0.09\linewidth]{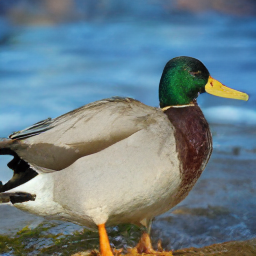} &
        \includegraphics[width=0.09\linewidth]{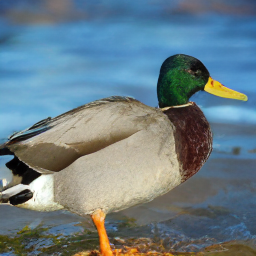} \\
        [0.2cm]

        \vspace{-1.6cm}
\rotatebox{90}{%
          \scriptsize
          \begin{tabular}{@{}l@{}}
            \multirow{2}{*}{\ditxltwo} \\
            
          \end{tabular}%
        } & & & & & & & & \\

         &
        \includegraphics[width=0.09\linewidth]{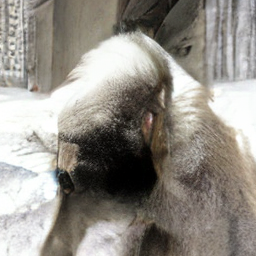} &
        \includegraphics[width=0.09\linewidth]{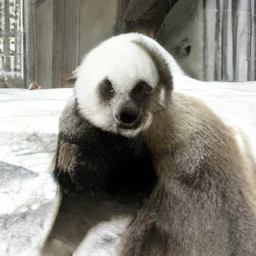} &
        \includegraphics[width=0.09\linewidth]{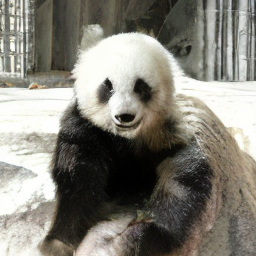} &
        \includegraphics[width=0.09\linewidth]{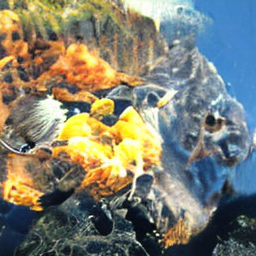} &
        \includegraphics[width=0.09\linewidth]{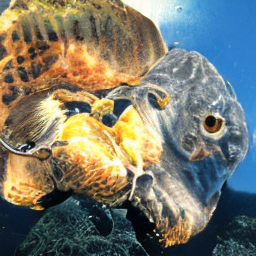} &
        \includegraphics[width=0.09\linewidth]{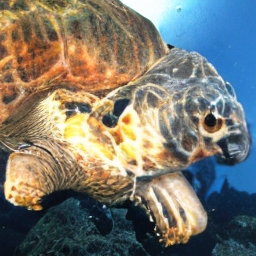} &
        \includegraphics[width=0.09\linewidth]{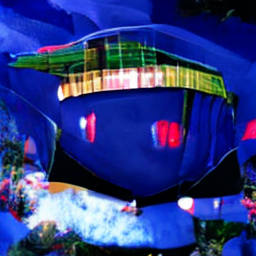} &
        \includegraphics[width=0.09\linewidth]{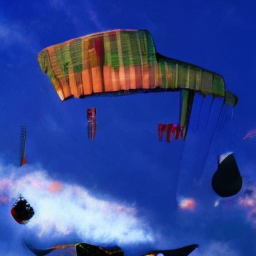} &
        \includegraphics[width=0.09\linewidth]{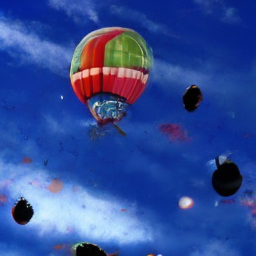} 
        \\[0.2cm]
        
        \vspace{-1.6cm}
        \rotatebox{90}{%
          \scriptsize
          \begin{tabular}{@{}l@{}}
            \ditxltwo \\
            w/ \our 
          \end{tabular}%
        } & & & & & & & & \\ 
        
        &
        \includegraphics[width=0.09\linewidth]{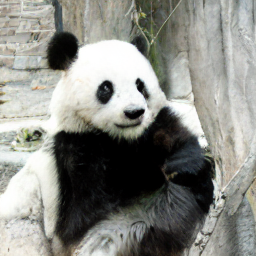} &
        \includegraphics[width=0.09\linewidth]{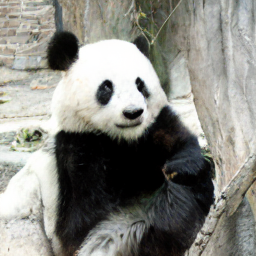} &
        \includegraphics[width=0.09\linewidth]{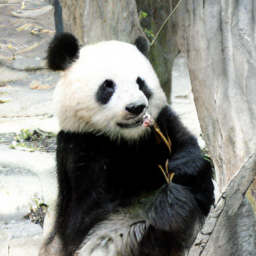} &
        \includegraphics[width=0.09\linewidth]{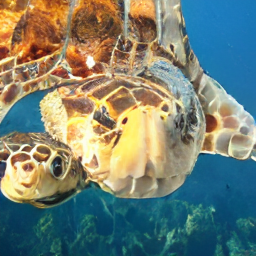} &
        \includegraphics[width=0.09\linewidth]{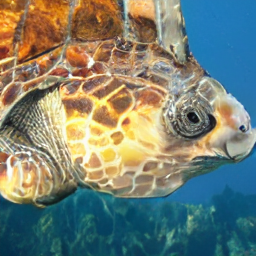} &
        \includegraphics[width=0.09\linewidth]{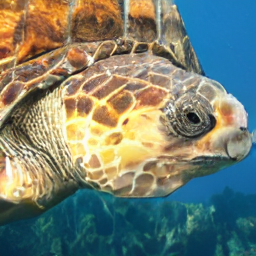} &
        \includegraphics[width=0.09\linewidth]{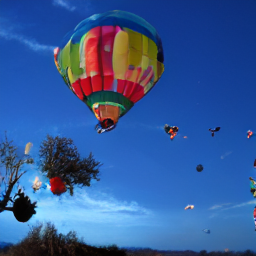} &
        \includegraphics[width=0.09\linewidth]{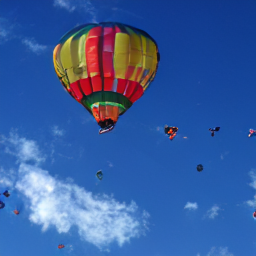} &
        \includegraphics[width=0.09\linewidth]{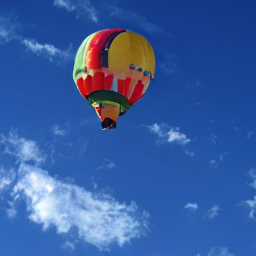} \\
        [2pt]

    \end{tabular}
    \vspace{-5pt}
    \caption{\textbf{\our accelerates generative modeling.} We compare results from two \ditxltwo models at 50K, 100K, and 400K iterations, one trained with \texttt{SD-VAE-FT-EMA} \textbf{(top)} and with \our \textbf{(bottom)}. 
    The same noise and number of sampling steps are used for both models, without classifier-free guidance. 
    Our approach delivers faster improvements in image quality, demonstrating accelerated convergence.}
    \label{tab:speedup}
    \vspace{-3pt}
\end{figure*}

\subsection{Boosting Generative Image Models}

By applying \our to both continuous and discrete autoencoders, we enhance the performance of state-of-the-art generative models, including \texttt{DiT} a pure transformer diffusion model, \texttt{SiT} that employs continuous flow-based modeling, \texttt{REPA} a recent approach aligning transformer representations with self-supervised features and \texttt{MaskGIT} a well-established masked generative model.
\vspace{-10pt}
\paragraph{DiT \& SiT} 
As demonstrated in ~\autoref{tab:bench-diff-main}, our regularization approach yields significant improvements across both \texttt{DiT-B} and \texttt{DiT-XL} models. Specifically, training \texttt{DiT-XL/2} on the regularized latent distribution achieves  \mbox{\Th{gFID} $14.5$} at $400\text{K}$ iterations, compared to \mbox{ $19.5$} without regularization. Notably, by $1.5\text{M}$ iterations, \texttt{DiT-XL/2} trained with \texttt{EQ-VAE} achieves \mbox{\Th{gFID} $8.8$}, outperforming the $\texttt{DiT-XL/2}$ model trained with \texttt{SD-VAE-FT-EMA} even at $7\text{M}$ iterations. The speed-up provided by \texttt{EQ-VAE} can be qualitatively observed in \autoref{tab:speedup}.
Moreover, in \autoref{tab:bench-diff-main}, we show that \texttt{SiT} models can also benefit from the regularized latent distribution of \texttt{EQ-VAE}, improving \Th{gFID} from $17.2$ to $16.1$ at $400\text{K}$ steps. 

\vspace{-10pt}
 \paragraph{REPA} We show that our regularization (which is performed in the first stage of latent generative modeling) is complementary to \texttt{REPA}, thus leading to further improvements in convergence and generation performance.
Specifically, training \texttt{REPA} (\texttt{SiT-XL-2}) with our \texttt{EQ-VAE} reaches  $5.9$ \Th{gFID} in $1\text{M}$ instead of $4\text{M}$ iterations. Thus, the regularized latent distribution of \texttt{EQ-VAE} can make the convergence of \texttt{REPA} $\times 4$ faster (\autoref{fig:teaser}). \emph{This is striking because \texttt{REPA} was shown to already significantly speed-up the convergence of diffusion models.}
\vspace{-10pt}
\paragraph{MaskGIT}
As shown in \autoref{tab:maskgit}, \texttt{MaskGIT} trained with our \our converges twice as fast reaching $6.80$ \Th{gFID} in $130$ epochs, instead of $300$. Furthermore, by epoch $300$ it reaches  $5.91$ \Th{gFID} surpassing the performance reported in both \cite{besnier2023pytorch} and \cite{chang2022maskgit}. 

\vspace{-3pt}
\paragraph{Comparison with state-of-the-art generative models}
\mbox{To further} demonstrate how \our accelerates the learning process, we compare it with recent diffusion methods using classifier-free guidance. Notably, as shown in \autoref{tab:bench-cfg-dif}, \ditxltwo with \our reaches $2.37$ \Th{gFID} in just $300$ epochs, matching the performance of \ditxltwo trained with \texttt{SD-VAE} or \texttt{SD-VAE-FT-MAE}. Even when combining  \our with the state-of-the-art approach \texttt{REPA}, we are able to achieve comparable results with standard \texttt{REPA} while using  $\times 4$ less training compute ($200$ vs $800$ epochs).

\begin{table}[t]
\footnotesize
\centering
\setlength{\tabcolsep}{1pt}
\begin{tabular}{l c c c c c c}
\toprule
\Th{Model} & \Th{Epochs}  &  { \Th{gFID}$\downarrow$} & {\Th{sFID}$\downarrow$} & {IS$\uparrow$} & {\Th{Pre.}$\uparrow$} & \Th{Rec.}$\uparrow$ \\
\arrayrulecolor{black}\midrule

 \texttt{LDM} & ~~200  & 3.60 & - & 247.7 & {0.87} & 0.48 \\

 \texttt{MaskDiT} & $1600$ &  $2.28$ & $5.67$ & $276.6$ & $0.80$ & $0.61$ \\ 
 \texttt{SD-DiT} & ~~$480$ & $3.23$ & -    & -     & -    & -     \\

 \sitxltwo   & $1400$ &     $2.06$ & $4.50$ & $270.3$ & $0.82$ & $0.59$ \\

  \midrule
   \ditxltwo   & $1400$  &    $2.27$ & $4.60$ & {$278.2$} & $0.83$ & $0.57$  \\
   
 \ditxltwo$^\dagger$   & $1400$  &    $2.47$ & $5.18$ & $276.1$ & $0.82$ & $0.57$  \\
  \rowcolor{TableColor} + \our (ours) & ~~{$300$} & {$2.37$} & {$4.78$} & {$277.3$} & {$0.82$} & {$0.57$} \\
 \midrule
\texttt{REPA}* & ~~{$800$} & $1.42$ & $4.70$ & $305.7$ & {$0.80$} & $0.65$ \\

\rowcolor{TableColor} + \our* (ours) & ~~{$200$} & {$1.70$} & {$5.13$} & {$283.0$} & {$0.79$} & {$0.62$} \\
\arrayrulecolor{black}\bottomrule
\end{tabular}
\vspace{-3pt}
\caption{\textbf{Comparison on ImageNet 256$\times$256 with CFG}.
\mbox{$\dagger$ indicates} that the used autoencoder is the original \texttt{SD-VAE} (instead of \texttt{SD-VAE-FT-EMA}). \texttt{REPA} uses \texttt{SiT-XL/2}.~* denotes that guidance interval~\citep{Kynkaanniemi2024} is applied.}
\label{tab:bench-cfg-dif}
\vspace{-10pt}
\end{table}

\subsection{Analysis}
\paragraph{Spatial transformations ablation} We begin the analysis of our method by ablating the effect of our equivariance regularization on generative performance with each spatial transformation to understand their respective impact. We consider isotropic $S(s,s)$ or anisotropic $S(s_x, s_y)$  scaling, rotations $R(\theta)$, and combined transformations. 
We then train a \texttt{DiT-B/2} on each latent distribution. In \autoref{tab:ablation-trans}, 
we observe that encouraging scale equivariance has a significant impact on generative performance.
Furthermore, rotation equivariance is also beneficial in generation performance.
Combining transformations yields further improvement, demonstrating their complementary effects. While anisotropic scaling yields a better generative performance since the regularization is more aggressive, it negatively impacts reconstruction quality. 
Thus, our \our default setting uses combinations of rotations and isotropic scaling.

\begin{table}[t]
\footnotesize
\centering
\setlength{\tabcolsep}{2.5pt}
\begin{tabular}{lcccc}
\toprule
\Th{Autoencoder} & $\tau$ & \Th{gFID$\downarrow$} & \Th{rFID$\downarrow$} & \Th{ID} \\ 
\midrule 
\sdvae & - & $43.5$ & $0.90$ & $62.2$ \\ \cmidrule(lr){1-5}
+ \texttt{EQ-VAE}  & $R(\theta)$ & $41.2$ & $0.73$ & $57.9$ \\
+ \texttt{EQ-VAE} & $S(s,s)$ & $35.8$ & $0.78$& $41.0$ \\
\rowcolor{TableColor} 
+ \texttt{EQ-VAE} & $R(\theta) \cdot S(s,s)$ & $34.1$ & $0.82$ &$39.4$ \\
+ \texttt{EQ-VAE} & $R(\theta) \cdot S(s_x,s_y)$ & $33.2$ &$0.92$& $38.9$ \\  
\bottomrule
\end{tabular}%
\vspace{-3pt}
\caption{\textbf{Spatial Transformation Ablation in \texttt{EQ-VAE}.}
We measure \Th{gFID}, \Th{rFID}, and intrinsic dimension (ID) for latents regularized via rotations, isotropic scaling, anisotropic scaling, and combinations. Combining transformations lowers ID and enhances generative performance, though anisotropic scaling can slightly degrade reconstruction.}
\label{tab:ablation-trans}
\end{table}

\begin{figure}[tbp]
    \centering
    \begin{tikzpicture}
      \begin{axis}[
        width=0.8\linewidth,
        height=0.550\linewidth,
        xmin=-0.5, xmax=5.5,
        ymin=33,  ymax=45,
        xtick={0,1,2,3,4,5},
        xticklabels={0,1,2,3,4,5},
        xlabel={\textbf{\# Finetuning Epochs w/ \texttt{EQ-VAE}}},
        ylabel={\textbf{\Th{gFID}-50K}},
        legend style={draw=none, at={(0.52,0.97)}, anchor=north west},
        grid=major,  
        major grid style={gray!20},
      ]

      \addplot [
        scatter,
        mesh,
        mark=*,
        line width=1.2pt,
        scatter src=x,   
        shader=interp,
        colormap name=myfade 
      ]
      coordinates {
        (0,43.5) (1,36.7) (2,35.9) (3,34.9) (4,34.2) (5,34.0)
      };

      \draw (axis cs:1.8,42.5) node[above left] {\textcolor[RGB]{44,62,80}{\texttt{\textbf{SD-VAE}}}};
      \draw (axis cs:5.5,35.5) node[above left] {\textcolor[RGB]{52,152,219}{\texttt{\textbf{+EQ-VAE}}}};
      \draw (axis cs:5.5,34.1) node[above left] {\textcolor[RGB]{52,152,219}{\texttt{(ours)}}};

      \end{axis}
    \end{tikzpicture}
\caption{\textbf{Rapid Improvement via \texttt{EQ-VAE} Fine-tuning.}  
Even a single epoch of \texttt{EQ-VAE} fine-tuning significantly improves generative modeling performance, reducing \Th{gFID} from 43.5 to 36.7. Generative modeling with \ditbtwo.}
\label{fig:ablation_epochs}
\vspace{-10pt}
\end{figure}
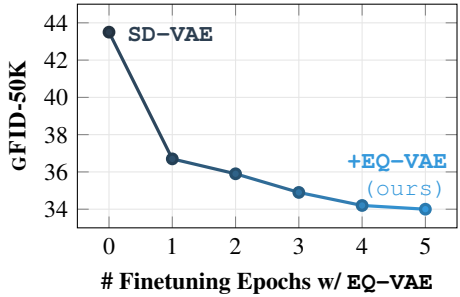

\vspace{-10pt}
\paragraph{Latent space complexity and generative performance}

To better understand the impact of our regularization on the complexity of the latent manifold, we measure its Intrinsic Dimension (ID). The ID represents the minimum number of variables needed to describe a data distribution \cite{1054365}. 
Notably, in \autoref{tab:ablation-trans}, we observe a correlation between the intrinsic dimension of the latent manifold and the resulting generative performance.
This suggests that the regularized latent distribution becomes simpler to model, further validating the effectiveness of our approach. This reduction in the complexity of latent representations can also be qualitatively observed in \autoref{fig:teaser} (left).
For further details on ID, see \autoref{sec:appendix_id}.


\paragraph{How many epochs does \texttt{EQ-VAE} need to enhance generation?} To demonstrate how quickly our objective regularizes the latent distribution, we conduct an ablation study by varying the number of fine-tuning epochs. We train a \ditbtwo model on the resulting latent distribution of each epoch and present the results in \autoref{fig:ablation_epochs}. Notably, even with a single epoch (10K steps) of fine-tuning, the \Th{gFID} drops from 43.5 to 36.7, highlighting the rapid refinement our objective achieves. For context, \texttt{SD-VAE-FT-EMA} has been fine-tuned for 300K steps.

\begin{table}[t]
\footnotesize
\centering
\setlength{\tabcolsep}{0.8pt}
\begin{tabular}{lcc}
\toprule
\Th{Autoencoder} & \Th{gFID $\downarrow$} & \Th{rFID $\downarrow$}   \\
\midrule 
 \texttt{SD-VAE} \cite{rombach2022high} & 43.8 & 0.90 \\ \hline
 \texttt{SD-VAE-FT-EMA} \cite{rombach2022high} & 43.5 & 0.73\\
\texttt{SD-VAE}$^\dagger$   & 43.5 & 0.81 \\
\texttt{EQ-VAE}  & 34.1 & 0.82 \\

\bottomrule
\end{tabular}
\vspace{-3pt}
\caption{\textbf{Additional Training vs. Equivariance Regularization.}  
Comparing various fine-tuning strategies for \texttt{SD-VAE} confirms that \our’s improvements stem from equivariance regularization. $^\dagger$ Denotes additional training with the standard objective (\Eq{eq:ldm}) for 5 epochs.}
\label{tab:ablation_ft}
\vspace{-3pt}
\end{table}

\paragraph{The enhancement in generative performance is not a result of the additional training} 
To verify that the improvement in generative performance stems from our equivariance regularization (\Eq{eq:ours_obj}) rather than additional training, we compare \texttt{EQ-VAE} with \texttt{SD-VAE}$^\dagger$ in \autoref{tab:ablation_ft}. \texttt{SD-VAE}$^\dagger$ is obtained by fine-tuning \texttt{SD-VAE} for five extra epochs using only the original objective (\Eq{eq:ldm}). The results show that this additional training has a negligible effect on generative performance, whereas \texttt{EQ-VAE} leads to a significant improvement. Similarly, \texttt{SD-VAE-EMA-FT}, derived from \texttt{SD-VAE}, has minimal impact on the \Th{gFID} score, further underscoring the effectiveness of \texttt{EQ-VAE}.

\section{Conclusion}
\label{sec:conclusion}
In this work, we argue that the structure of latent representations produced by the autoencoder is crucial for the convergence speed and performance of latent generative models.
We observed that latent representations of established autoencoders are not equivariant under simple spatial transformations. To address this, we introduce \texttt{EQ-VAE}, a simple modification to the autoencoder's training objective. We empirically demonstrated that fine-tuning pre-trained autoencoders with \texttt{EQ-VAE} for just a few epochs, is enough to reduce the equivariance error and significantly boost the performance of latent generative models while maintaining their reconstruction capability. We believe that our work introduces several promising future directions, particularly in exploring the theoretical and empirical relationship between the geometry of the latent distribution and the performance of latent generative models.
\vspace{-0.1cm}
\section*{Impact Statement}
\vspace{-0.1cm}
This paper presents work whose goal is to advance the field
of machine learning in general and image synthesis in particular. There are many potential societal consequences
of our work, none of which we feel must be specifically
highlighted here.
\paragraph{Acknowledgements}

This work has been partially supported by project MIS 5154714 of the National Recovery and Resilience Plan Greece 2.0 funded by the European Union under the NextGenerationEU Program. 

Hardware resources were granted with the support of GRNET. Also, this work was performed using HPC resources from GENCI-IDRIS (Grants 2024-AD011012884R3).

\addtocontents{toc}{\protect\setcounter{tocdepth}{2}}

\bibliography{main}

\begin{thebibliography}{74}
\providecommand{\natexlab}[1]{#1}
\providecommand{\url}[1]{\texttt{#1}}
\expandafter\ifx\csname urlstyle\endcsname\relax
  \providecommand{\doi}[1]{doi: #1}\else
  \providecommand{\doi}{doi: \begingroup \urlstyle{rm}\Url}\fi

\bibitem[Bennett(1969)]{1054365}
Bennett, R.
\newblock The intrinsic dimensionality of signal collections.
\newblock \emph{IEEE Transactions on Information Theory}, 15\penalty0 (5):\penalty0 517--525, 1969.
\newblock \doi{10.1109/TIT.1969.1054365}.

\bibitem[Besnier \& Chen(2023)Besnier and Chen]{besnier2023pytorch}
Besnier, V. and Chen, M.
\newblock A pytorch reproduction of masked generative image transformer.
\newblock \emph{arXiv preprint arXiv:2310.14400}, 2023.

\bibitem[{Black Forest Labs}(2023)]{flux2023}
{Black Forest Labs}.
\newblock Flux, 2023.

\bibitem[Brehmer et~al.(2024)Brehmer, Behrends, de~Haan, and Cohen]{brehmer2024does}
Brehmer, J., Behrends, S., de~Haan, P., and Cohen, T.
\newblock Does equivariance matter at scale?
\newblock \emph{arXiv preprint arXiv:2410.23179}, 2024.

\bibitem[Chang et~al.(2022)Chang, Zhang, Jiang, Liu, and Freeman]{chang2022maskgit}
Chang, H., Zhang, H., Jiang, L., Liu, C., and Freeman, W.~T.
\newblock Maskgit: Masked generative image transformer.
\newblock In \emph{CVPR}, pp.\  11315--11325, 2022.

\bibitem[Chen et~al.(2024)Chen, YU, GE, Yao, Xie, Wang, Kwok, Luo, Lu, and Li]{chen2024pixartalpha}
Chen, J., YU, J., GE, C., Yao, L., Xie, E., Wang, Z., Kwok, J., Luo, P., Lu, H., and Li, Z.
\newblock Pixart-\${\textbackslash}alpha\$: Fast training of diffusion transformer for photorealistic text-to-image synthesis.
\newblock In \emph{ICLR}, 2024.

\bibitem[Chen et~al.(2025)Chen, Cai, Chen, Xie, Yang, Tang, Li, Lu, and Han]{chen2025deep}
Chen, J., Cai, H., Chen, J., Xie, E., Yang, S., Tang, H., Li, M., Lu, Y., and Han, S.
\newblock Deep compression autoencoder for efficient high-resolution diffusion models.
\newblock In \emph{ICLR}, 2025.

\bibitem[Cheng et~al.(2023)Cheng, Kervadec, and Baroni]{cheng2023bridging}
Cheng, E., Kervadec, C., and Baroni, M.
\newblock Bridging information-theoretic and geometric compression in language models.
\newblock \emph{arXiv preprint arXiv:2310.13620}, 2023.

\bibitem[Cohen \& Welling(2016)Cohen and Welling]{cohen2016group}
Cohen, T. and Welling, M.
\newblock Group equivariant convolutional networks.
\newblock In \emph{ICLR}, pp.\  2990--2999. PMLR, 2016.

\bibitem[Dai et~al.(2023)Dai, Hou, Ma, Tsai, Wang, Wang, Zhang, Vandenhende, Wang, Dubey, Yu, Kadian, Radenovic, Mahajan, Li, Zhao, Petrovic, Singh, Motwani, and Wen]{dai2023emu}
Dai, X., Hou, J., Ma, C.-Y., Tsai, S., Wang, J., Wang, R., Zhang, P., Vandenhende, S., Wang, X., Dubey, A., Yu, M., Kadian, A., Radenovic, F., Mahajan, D., Li, K., Zhao, Y., Petrovic, V., Singh, M.~K., Motwani, S., and Wen, Y.
\newblock Emu: Enhancing image generation models using photogenic needles in a haystack.
\newblock \emph{arXiv preprint arXiv:2309.15807}, 2023.

\bibitem[Deng et~al.(2009)Deng, Dong, Socher, Li, Li, and Fei-Fei]{deng2009imagenet}
Deng, J., Dong, W., Socher, R., Li, L.-J., Li, K., and Fei-Fei, L.
\newblock Imagenet: A large-scale hierarchical image database.
\newblock In \emph{CVPR}, pp.\  248--255, 2009.

\bibitem[Dhariwal \& Nichol(2021)Dhariwal and Nichol]{dhariwal2021adm}
Dhariwal, P. and Nichol, A.~Q.
\newblock Diffusion models beat {GAN}s on image synthesis.
\newblock In \emph{NeurIPS}, 2021.

\bibitem[Dilokthanakul et~al.(2016)Dilokthanakul, Mediano, Garnelo, Lee, Salimbeni, Arulkumaran, and Shanahan]{dilokthanakul2016deep}
Dilokthanakul, N., Mediano, P.~A., Garnelo, M., Lee, M.~C., Salimbeni, H., Arulkumaran, K., and Shanahan, M.
\newblock Deep unsupervised clustering with gaussian mixture variational autoencoders.
\newblock \emph{arXiv preprint arXiv:1611.02648}, 2016.

\bibitem[Esser et~al.(2021)Esser, Rombach, and Ommer]{esser2021taming}
Esser, P., Rombach, R., and Ommer, B.
\newblock Taming transformers for high-resolution image synthesis.
\newblock In \emph{CVPR}, pp.\  4195--4205, 2021.

\bibitem[Esser et~al.(2024)Esser, Kulal, Blattmann, Entezari, M\"{u}ller, Saini, Levi, Lorenz, Sauer, Boesel, Podell, Dockhorn, English, and Rombach]{esser2024sd3}
Esser, P., Kulal, S., Blattmann, A., Entezari, R., M\"{u}ller, J., Saini, H., Levi, Y., Lorenz, D., Sauer, A., Boesel, F., Podell, D., Dockhorn, T., English, Z., and Rombach, R.
\newblock Scaling rectified flow transformers for high-resolution image synthesis.
\newblock In \emph{ICML}, pp.\  12606--12633, 2024.

\bibitem[Facco et~al.(2017)Facco, d’Errico, Rodriguez, and Laio]{facco2017estimating}
Facco, E., d’Errico, M., Rodriguez, A., and Laio, A.
\newblock Estimating the intrinsic dimension of datasets by a minimal neighborhood information.
\newblock \emph{Scientific reports}, 7\penalty0 (1):\penalty0 12140, 2017.

\bibitem[Glielmo et~al.(2022)Glielmo, Macocco, Doimo, Carli, Zeni, Wild, d’Errico, Rodriguez, and Laio]{glielmo2022dadapy}
Glielmo, A., Macocco, I., Doimo, D., Carli, M., Zeni, C., Wild, R., d’Errico, M., Rodriguez, A., and Laio, A.
\newblock Dadapy: Distance-based analysis of data-manifolds in python.
\newblock \emph{Patterns}, 3\penalty0 (10), 2022.

\bibitem[Heusel et~al.(2017)Heusel, Ramsauer, Unterthiner, Nessler, and Hochreiter]{fid}
Heusel, M., Ramsauer, H., Unterthiner, T., Nessler, B., and Hochreiter, S.
\newblock Gans trained by a two time-scale update rule converge to a local nash equilibrium.
\newblock \emph{Advances in neural information processing systems}, 30, 2017.

\bibitem[Higgins et~al.(2016)Higgins, Matthey, Pal, Burgess, Glorot, Botvinick, Mohamed, and Lerchner]{Higgins2016betaVAELB}
Higgins, I., Matthey, L., Pal, A., Burgess, C.~P., Glorot, X., Botvinick, M.~M., Mohamed, S., and Lerchner, A.
\newblock beta-vae: Learning basic visual concepts with a constrained variational framework.
\newblock In \emph{ICLR}, 2016.

\bibitem[Hoogeboom et~al.(2018)Hoogeboom, Peters, Cohen, and Welling]{hoogeboom2018hexaconv}
Hoogeboom, E., Peters, J.~W., Cohen, T.~S., and Welling, M.
\newblock Hexaconv.
\newblock \emph{arXiv preprint arXiv:1803.02108}, 2018.

\bibitem[Hu et~al.(2023)Hu, Chen, Wang, Li, Wang, Sun, and Li]{hu2023complexity}
Hu, T., Chen, F., Wang, H., Li, J., Wang, W., Sun, J., and Li, Z.
\newblock Complexity matters: Rethinking the latent space for generative modeling.
\newblock In \emph{NeurIPS}, 2023.

\bibitem[Isola et~al.(2017)Isola, Zhu, Zhou, and Efros]{isola2017image}
Isola, P., Zhu, J.-Y., Zhou, T., and Efros, A.~A.
\newblock Image-to-image translation with conditional adversarial networks.
\newblock In \emph{CVPR}, pp.\  1125--1134, 2017.

\bibitem[Kaba et~al.(2023)Kaba, Mondal, Zhang, Bengio, and Ravanbakhsh]{kaba2023equivariance}
Kaba, S.-O., Mondal, A.~K., Zhang, Y., Bengio, Y., and Ravanbakhsh, S.
\newblock Equivariance with learned canonicalization functions.
\newblock In \emph{International Conference on Machine Learning}, pp.\  15546--15566. PMLR, 2023.

\bibitem[Kingma \& Welling(2014)Kingma and Welling]{kingma2014}
Kingma, D.~P. and Welling, M.
\newblock Auto-encoding variational bayes.
\newblock In \emph{ICLR}, 2014.

\bibitem[Kolouri et~al.(2018)Kolouri, Pope, Martin, and Rohde]{kolouri2018sliced}
Kolouri, S., Pope, P.~E., Martin, C.~E., and Rohde, G.~K.
\newblock Sliced wasserstein auto-encoders.
\newblock In \emph{ICLR}, 2018.

\bibitem[Kondor(2018)]{kondor2018n}
Kondor, R.
\newblock N-body networks: a covariant hierarchical neural network architecture for learning atomic potentials.
\newblock \emph{arXiv preprint arXiv:1803.01588}, 2018.

\bibitem[Kvinge et~al.(2023)Kvinge, Brown, and Godfrey]{kvinge2023exploring}
Kvinge, H., Brown, D., and Godfrey, C.
\newblock Exploring the representation manifolds of stable diffusion through the lens of intrinsic dimension.
\newblock \emph{arXiv preprint arXiv:2302.09301}, 2023.

\bibitem[Kynk{\"a}{\"a}nniemi et~al.(2019)Kynk{\"a}{\"a}nniemi, Karras, Laine, Lehtinen, and Aila]{kynkaanniemi2019improved}
Kynk{\"a}{\"a}nniemi, T., Karras, T., Laine, S., Lehtinen, J., and Aila, T.
\newblock Improved precision and recall metric for assessing generative models.
\newblock \emph{Advances in neural information processing systems}, 32, 2019.

\bibitem[Kynkäänniemi et~al.(2024)Kynkäänniemi, Aittala, Karras, Laine, Aila, and Lehtinen]{Kynkaanniemi2024}
Kynkäänniemi, T., Aittala, M., Karras, T., Laine, S., Aila, T., and Lehtinen, J.
\newblock Applying guidance in a limited interval improves sample and distribution quality in diffusion models.
\newblock In \emph{NeurIPS}, 2024.

\bibitem[Li et~al.(2023)Li, Chang, Mishra, Zhang, Katabi, and Krishnan]{li2023mage}
Li, T., Chang, H., Mishra, S., Zhang, H., Katabi, D., and Krishnan, D.
\newblock Mage: Masked generative encoder to unify representation learning and image synthesis.
\newblock In \emph{CVPR}, pp.\  2142--2152, 2023.

\bibitem[Li et~al.(2024)Li, Tian, Li, Deng, and He]{li2024autoregressive}
Li, T., Tian, Y., Li, H., Deng, M., and He, K.
\newblock Autoregressive image generation without vector quantization.
\newblock \emph{arXiv preprint arXiv:2406.11838}, 2024.

\bibitem[Ma et~al.(2024)Ma, Goldstein, Albergo, Boffi, Vanden-Eijnden, and Xie]{ma2024sit}
Ma, N., Goldstein, M., Albergo, M.~S., Boffi, N.~M., Vanden-Eijnden, E., and Xie, S.
\newblock Sit: Exploring flow and diffusion-based generative models with scalable interpolant transformers.
\newblock In \emph{ECCV}, pp.\  23–40, 2024.

\bibitem[Makhzani et~al.(2015)Makhzani, Shlens, Jaitly, Goodfellow, and Frey]{makhzani2015adversarial}
Makhzani, A., Shlens, J., Jaitly, N., Goodfellow, I., and Frey, B.
\newblock Adversarial autoencoders.
\newblock \emph{arXiv preprint arXiv:1511.05644}, 2015.

\bibitem[Marcos et~al.(2017)Marcos, Volpi, Komodakis, and Tuia]{marcos2017rotation}
Marcos, D., Volpi, M., Komodakis, N., and Tuia, D.
\newblock Rotation equivariant vector field networks.
\newblock In \emph{Proceedings of the IEEE International Conference on Computer Vision}, pp.\  5048--5057, 2017.

\bibitem[Nash et~al.(2021)Nash, Menick, Dieleman, and Battaglia]{sfid}
Nash, C., Menick, J., Dieleman, S., and Battaglia, P.~W.
\newblock Generating images with sparse representations.
\newblock \emph{arXiv preprint arXiv:2103.03841}, 2021.

\bibitem[Nichol \& Dhariwal(2021)Nichol and Dhariwal]{nichol21a}
Nichol, A.~Q. and Dhariwal, P.
\newblock Improved denoising diffusion probabilistic models.
\newblock In \emph{ICML}, volume 139, pp.\  8162--8171, 18--24 Jul 2021.

\bibitem[Peebles \& Xie(2023)Peebles and Xie]{peebles2023scalable}
Peebles, W. and Xie, S.
\newblock Scalable diffusion models with transformers.
\newblock In \emph{Proceedings of the IEEE/CVF International Conference on Computer Vision}, pp.\  4195--4205, 2023.

\bibitem[Podell et~al.(2024)Podell, English, Lacey, Blattmann, Dockhorn, M{\"u}ller, Penna, and Rombach]{podell2024sdxl}
Podell, D., English, Z., Lacey, K., Blattmann, A., Dockhorn, T., M{\"u}ller, J., Penna, J., and Rombach, R.
\newblock {SDXL}: Improving latent diffusion models for high-resolution image synthesis.
\newblock In \emph{ICLR}, 2024.

\bibitem[Pope et~al.(2021)Pope, Zhu, Abdelkader, Goldblum, and Goldstein]{pope2021the}
Pope, P., Zhu, C., Abdelkader, A., Goldblum, M., and Goldstein, T.
\newblock The intrinsic dimension of images and its impact on learning.
\newblock In \emph{ICLR}, 2021.

\bibitem[Puny et~al.(2022)Puny, Atzmon, Smith, Misra, Grover, Ben-Hamu, and Lipman]{puny2022frame}
Puny, O., Atzmon, M., Smith, E.~J., Misra, I., Grover, A., Ben-Hamu, H., and Lipman, Y.
\newblock Frame averaging for invariant and equivariant network design.
\newblock In \emph{ICLR}, 2022.

\bibitem[Rahman \& Yeh(2023)Rahman and Yeh]{rahman2023truly}
Rahman, M.~A. and Yeh, R.~A.
\newblock Truly scale-equivariant deep nets with fourier layers.
\newblock \emph{Advances in Neural Information Processing Systems}, 36:\penalty0 6092--6104, 2023.

\bibitem[Ravanbakhsh et~al.(2017)Ravanbakhsh, Schneider, and Poczos]{ravanbakhsh2017equivariance}
Ravanbakhsh, S., Schneider, J., and Poczos, B.
\newblock Equivariance through parameter-sharing.
\newblock In \emph{International conference on machine learning}, pp.\  2892--2901. PMLR, 2017.

\bibitem[Rombach et~al.(2022)Rombach, Blattmann, Lorenz, Esser, and Ommer]{rombach2022high}
Rombach, R., Blattmann, A., Lorenz, D., Esser, P., and Ommer, B.
\newblock High-resolution image synthesis with latent diffusion models.
\newblock In \emph{CVPR}, pp.\  10684--10695, 2022.

\bibitem[Ryu(2024)]{ryu2024vqgan}
Ryu, S.
\newblock Training vqgan and vae, with detailed explanation.
\newblock \url{https://github.com/cloneofsimo/vqgan-training}, 2024.
\newblock GitHub repository.

\bibitem[Salimans et~al.(2016)Salimans, Goodfellow, Zaremba, Cheung, Radford, and Chen]{is}
Salimans, T., Goodfellow, I., Zaremba, W., Cheung, V., Radford, A., and Chen, X.
\newblock Improved techniques for training gans.
\newblock \emph{Advances in neural information processing systems}, 29, 2016.

\bibitem[Schuhmann et~al.(2022)Schuhmann, Beaumont, Vencu, Gordon, Wightman, Cherti, Coombes, Katta, Mullis, Wortsman, Schramowski, Kundurthy, Crowson, Schmidt, Kaczmarczyk, and Jitsev]{schuhmann2022laion}
Schuhmann, C., Beaumont, R., Vencu, R., Gordon, C., Wightman, R., Cherti, M., Coombes, T., Katta, A., Mullis, C., Wortsman, M., Schramowski, P., Kundurthy, S., Crowson, K., Schmidt, L., Kaczmarczyk, R., and Jitsev, J.
\newblock Laion-5b: An open large-scale dataset for training next generation image-text models.
\newblock In \emph{NeurIPS}, volume~35, pp.\  25278--25294, 2022.

\bibitem[Simonyan(2014)]{vgg}
Simonyan, K.
\newblock Very deep convolutional networks for large-scale image recognition.
\newblock \emph{arXiv preprint arXiv:1409.1556}, 2014.

\bibitem[Sinha \& Dieng(2021)Sinha and Dieng]{NEURIPS2021_6c19e0a6}
Sinha, S. and Dieng, A.~B.
\newblock Consistency regularization for variational auto-encoders.
\newblock In Ranzato, M., Beygelzimer, A., Dauphin, Y., Liang, P., and Vaughan, J.~W. (eds.), \emph{Advances in Neural Information Processing Systems}, volume~34, pp.\  12943--12954. Curran Associates, Inc., 2021.

\bibitem[Sosnovik et~al.(2020)Sosnovik, Szmaja, and Smeulders]{Sosnovik2020Scale-Equivariant}
Sosnovik, I., Szmaja, M., and Smeulders, A.
\newblock Scale-equivariant steerable networks.
\newblock In \emph{ICLR}, 2020.

\bibitem[Szegedy et~al.(2016)Szegedy, Vanhoucke, Ioffe, Shlens, and Wojna]{szegedy2016rethinking}
Szegedy, C., Vanhoucke, V., Ioffe, S., Shlens, J., and Wojna, Z.
\newblock Rethinking the inception architecture for computer vision.
\newblock In \emph{CVPR}, pp.\  2818--2826, 2016.

\bibitem[Takahashi et~al.(2019)Takahashi, Iwata, Yamanaka, Yamada, and Yagi]{takahashi2019variational}
Takahashi, H., Iwata, T., Yamanaka, Y., Yamada, M., and Yagi, S.
\newblock Variational autoencoder with implicit optimal priors.
\newblock In \emph{Proceedings of the AAAI Conference on Artificial Intelligence}, volume~33, pp.\  5066--5073, 2019.

\bibitem[Thomas et~al.(2018)Thomas, Smidt, Kearnes, Yang, Li, Kohlhoff, and Riley]{thomas2018tensor}
Thomas, N., Smidt, T., Kearnes, S., Yang, L., Li, L., Kohlhoff, K., and Riley, P.
\newblock Tensor field networks: Rotation-and translation-equivariant neural networks for 3d point clouds.
\newblock \emph{arXiv preprint arXiv:1802.08219}, 2018.

\bibitem[Tian et~al.(2024)Tian, Jiang, Yuan, Peng, and Wang]{tian2024visual}
Tian, K., Jiang, Y., Yuan, Z., Peng, B., and Wang, L.
\newblock Visual autoregressive modeling: Scalable image generation via next-scale prediction.
\newblock \emph{arXiv preprint arXiv:2404.02905}, 2024.

\bibitem[Tolstikhin et~al.(2018)Tolstikhin, Bousquet, Gelly, and Schoelkopf]{tolstikhin2018wasserstein}
Tolstikhin, I., Bousquet, O., Gelly, S., and Schoelkopf, B.
\newblock Wasserstein auto-encoders.
\newblock In \emph{ICLR}, 2018.

\bibitem[Tomczak \& Welling(2018)Tomczak and Welling]{tomczak2018vae}
Tomczak, J. and Welling, M.
\newblock Vae with a vampprior.
\newblock In \emph{International conference on artificial intelligence and statistics}, pp.\  1214--1223. PMLR, 2018.

\bibitem[Tschannen et~al.(2025)Tschannen, Eastwood, and Mentzer]{tschannen2025givt}
Tschannen, M., Eastwood, C., and Mentzer, F.
\newblock Givt: Generative infinite-vocabulary transformers.
\newblock In \emph{ECCV}, pp.\  292--309, 2025.

\bibitem[Valeriani et~al.(2023)Valeriani, Doimo, Cuturello, Laio, Ansuini, and Cazzaniga]{valeriani2023geometry}
Valeriani, L., Doimo, D., Cuturello, F., Laio, A., Ansuini, A., and Cazzaniga, A.
\newblock The geometry of hidden representations of large transformer models.
\newblock \emph{Advances in Neural Information Processing Systems}, 36:\penalty0 51234--51252, 2023.

\bibitem[van~den Oord et~al.(2017)van~den Oord, Vinyals, and kavukcuoglu]{oord2017vq}
van~den Oord, A., Vinyals, O., and kavukcuoglu, k.
\newblock Neural discrete representation learning.
\newblock In \emph{NeurIPS}, volume~30, 2017.

\bibitem[Wang et~al.(2004)Wang, Bovik, Sheikh, and Simoncelli]{ssim}
Wang, Z., Bovik, A., Sheikh, H., and Simoncelli, E.
\newblock Image quality assessment: from error visibility to structural similarity.
\newblock \emph{IEEE Transactions on Image Processing}, 13\penalty0 (4):\penalty0 600--612, 2004.
\newblock \doi{10.1109/TIP.2003.819861}.

\bibitem[Weiler \& Cesa(2019)Weiler and Cesa]{weiler2019general}
Weiler, M. and Cesa, G.
\newblock General e (2)-equivariant steerable cnns.
\newblock \emph{Advances in neural information processing systems}, 32, 2019.

\bibitem[Winter et~al.(2022)Winter, Bertolini, Le, Noe, and Clevert]{winter2022unsupervised}
Winter, R., Bertolini, M., Le, T., Noe, F., and Clevert, D.-A.
\newblock Unsupervised learning of group invariant and equivariant representations.
\newblock In \emph{NeurIPS}, 2022.

\bibitem[Worrall \& Brostow(2018)Worrall and Brostow]{worrall2018cubenet}
Worrall, D. and Brostow, G.
\newblock Cubenet: Equivariance to 3d rotation and translation.
\newblock In \emph{ECCV}, pp.\  567--584, 2018.

\bibitem[Xie et~al.(2025)Xie, Chen, Chen, Cai, Tang, Lin, Zhang, Li, Zhu, Lu, and Han]{xie2025sana}
Xie, E., Chen, J., Chen, J., Cai, H., Tang, H., Lin, Y., Zhang, Z., Li, M., Zhu, L., Lu, Y., and Han, S.
\newblock {SANA}: Efficient high-resolution text-to-image synthesis with linear diffusion transformers.
\newblock In \emph{ICLR}, 2025.

\bibitem[Yao et~al.(2024)Yao, Wang, Liu, and Wang]{yao2024fasterdit}
Yao, J., Wang, C., Liu, W., and Wang, X.
\newblock Fasterdit: Towards faster diffusion transformers training without architecture modification.
\newblock In \emph{NeurIPS}, 2024.

\bibitem[Yu et~al.(2022{\natexlab{a}})Yu, Li, Koh, Zhang, Pang, Qin, Ku, Xu, Baldridge, and Wu]{yu2022vectorquantized}
Yu, J., Li, X., Koh, J.~Y., Zhang, H., Pang, R., Qin, J., Ku, A., Xu, Y., Baldridge, J., and Wu, Y.
\newblock Vector-quantized image modeling with improved {VQGAN}.
\newblock In \emph{ICLR}, 2022{\natexlab{a}}.

\bibitem[Yu et~al.(2022{\natexlab{b}})Yu, Xu, Koh, Luong, Baid, Wang, Vasudevan, Ku, Yang, Ayan, et~al.]{yu2022scaling}
Yu, J., Xu, Y., Koh, J.~Y., Luong, T., Baid, G., Wang, Z., Vasudevan, V., Ku, A., Yang, Y., Ayan, B.~K., et~al.
\newblock Scaling autoregressive models for content-rich text-to-image generation.
\newblock \emph{arXiv preprint arXiv:2206.10789}, 2022{\natexlab{b}}.

\bibitem[Yu et~al.(2024)Yu, Lezama, Gundavarapu, Versari, Sohn, Minnen, Cheng, Gupta, Gu, Hauptmann, Gong, Yang, Essa, Ross, and Jiang]{yu2024language}
Yu, L., Lezama, J., Gundavarapu, N.~B., Versari, L., Sohn, K., Minnen, D., Cheng, Y., Gupta, A., Gu, X., Hauptmann, A.~G., Gong, B., Yang, M.-H., Essa, I., Ross, D.~A., and Jiang, L.
\newblock Language model beats diffusion - tokenizer is key to visual generation.
\newblock In \emph{The Twelfth International Conference on Learning Representations}, 2024.

\bibitem[Yu et~al.(2025)Yu, Kwak, Jang, Jeong, Huang, Shin, and Xie]{Yu2025repa}
Yu, S., Kwak, S., Jang, H., Jeong, J., Huang, J., Shin, J., and Xie, S.
\newblock Representation alignment for generation: Training diffusion transformers is easier than you think.
\newblock In \emph{ICLR}, 2025.

\bibitem[Zhang et~al.(2018)Zhang, Isola, Efros, Shechtman, and Wang]{zhang2018unreasonable}
Zhang, R., Isola, P., Efros, A.~A., Shechtman, E., and Wang, O.
\newblock The unreasonable effectiveness of deep features as a perceptual metric.
\newblock In \emph{CVPR}, pp.\  586--595, 2018.

\bibitem[Zhao et~al.(2018)Zhao, Kim, Zhang, Rush, and LeCun]{zhao2018adversa}
Zhao, J., Kim, Y., Zhang, K., Rush, A., and LeCun, Y.
\newblock Adversarially regularized autoencoders.
\newblock In \emph{ICML}, volume~80, pp.\  5902--5911, 2018.

\bibitem[Zheng et~al.(2023)Zheng, Nie, Vahdat, and Anandkumar]{zheng2023fast}
Zheng, H., Nie, W., Vahdat, A., and Anandkumar, A.
\newblock Fast training of diffusion models with masked transformers.
\newblock \emph{arXiv preprint arXiv:2306.09305}, 2023.

\bibitem[Zhu et~al.(2024{\natexlab{a}})Zhu, Wei, Lu, and Chen]{zhu2024scaling}
Zhu, L., Wei, F., Lu, Y., and Chen, D.
\newblock Scaling the codebook size of {VQ}-{GAN} to 100,000 with a utilization rate of 99\%.
\newblock In \emph{The Thirty-eighth Annual Conference on Neural Information Processing Systems}, 2024{\natexlab{a}}.

\bibitem[Zhu et~al.(2024{\natexlab{b}})Zhu, Pan, Li, Yao, Sun, Mei, and Chen]{zhu2024sd}
Zhu, R., Pan, Y., Li, Y., Yao, T., Sun, Z., Mei, T., and Chen, C.~W.
\newblock Sd-dit: Unleashing the power of self-supervised discrimination in diffusion transformer.
\newblock In \emph{CVPR}, pp.\  8435--8445, 2024{\natexlab{b}}.

\bibitem[Zhu et~al.(2023)Zhu, Feng, Chen, Bao, Wang, Chen, Yuan, and Hua]{zhu2023designing}
Zhu, Z., Feng, X., Chen, D., Bao, J., Wang, L., Chen, Y., Yuan, L., and Hua, G.
\newblock Designing a better asymmetric vqgan for stablediffusion.
\newblock \emph{arXiv preprint arXiv:2306.04632}, 2023.

\end{thebibliography}
\bibliographystyle{icml2025}

\clearpage
\appendix
\onecolumn

\renewcommand{\thesection}{\Alph{section}}
{
  \hypersetup{linkcolor=black}
  \tableofcontents
}

\section{Additional Ablations}
\label{more_ablations}

\subsection{Implicit vs Explicit Equivariance Regularization}
\label{sec:appenidx_exp_imp}

Here, we provide an analysis of the design choice of our objective. We aim to design an objective that reduces the equivariance error of the encoder while avoiding mode collapse and preserving reconstruction performance. For each objective investigated, we finetune \texttt{SD-VAE} and evaluate the effect on generative performance by training a \texttt{DiT-B/2} on the resulting latent distribution. 
Initially, we perform fine-tuning with the standard objective along with the explicit loss in (\Eq{eq:encoder_loss}):\mbox{ $\mathcal{L}_{VAE} +   \lambda  \mathcal{L}_{\text{explicit}}$ }and set $\lambda=0.1$. 
We further experiment with adding a stop-gradient (sg) in the $\mathcal{E}(\mathbf{\tau \circ x})$ term in $\mathcal{L}_{\text{explicit}}$.
In ~\autoref{tab:losses}, we observe that using $\mathcal{L}_{\text{explicit}}$ successfully reduces the equivariance error for both rotation and scaling transformations.  However, both reconstruction and generative performance degrade severely, indicating a mode collapse in the latent space. 

\begin{table}[!h]

\centering
\setlength{\tabcolsep}{3pt}
\begin{tabular}{lccccc}
\hline
\multirow{2}{*}{\Th{Loss}} & \multirow{2}{*}{\Th{gFID$\downarrow$}} 
& \multirow{2}{*}{\Th{rFID$\downarrow$}} & \multicolumn{2}{c}{\Th{Equivariance Error}} \\ \cmidrule(lr){4-5}
 &  &  & \hspace{0.5cm} $R(\theta) \downarrow$ & $S(s) \downarrow$ \\ \hline
\texttt{SD-VAE} & ~~$43.5$ & ~~~~$0.90$  &    \hspace{0.5cm} $0.93$      &   $0.80$       \\
 w/ explicit& $141.3$ & $117.93$   &    \hspace{0.5cm}  $0.32$      &   $0.11$       \\
 w/ explicit + sg& $134.7$ & $109.25$   &    \hspace{0.5cm}  $0.35$      &   $0.13$       \\
 
\rowcolor{TableColor}  w/ implicit (ours) & ~~$34.1$ & ~~~~$0.82$   &    \hspace{0.5cm}  $0.49$      &   $0.15$          \\ \hline
\end{tabular}
\caption{\textbf{Implicit vs. Explicit Equivariance Regularization.}  
Comparing \texttt{SD-VAE} along with explicit vs implicit regularization objectives shows that explicit regularization drastically lowers equivariance errors but triggers mode collapse, while implicit regularization enhances significantly the generative performance.}
\label{tab:losses}
\end{table}

\subsection{Regularization Strength}
We evaluate the impact of hyperparameter $p_{\alpha}$ which controls the strength of our regularization  in \autoref{tab:ablation_prior}. 
\label{sec:appendix_prior}
\begin{table}[!h]
\centering
\setlength{\tabcolsep}{2.5pt}
\begin{tabular}{lccc}
\toprule
\Th{Autoencoder} & $p_\alpha$ & \Th{gFID $\downarrow$} & \Th{rFID $\downarrow$}   \\
\midrule 
\texttt{EQ-VAE}  & 0.3 & 35.4 & 0.78 \\
\texttt{EQ-VAE}  & 0.7 & 34.4 & 0.88 \\
\rowcolor{TableColor} \texttt{EQ-VAE}  & 0.5 & 34.1 & 0.82 \\
\bottomrule
\end{tabular}
\vspace{-3pt}
\caption{\textbf{Ablation on regularization strength.}  We perform two experiments, with lower ($p_{\alpha} = 0.7$) and higher ($p_{\alpha}= 0.3$) regularization strength. We observe that our method is relatively robust to choices of $p_{\alpha}$. We highlight the setting used throughout all our experiments.}
\label{tab:ablation_prior}
\vspace{-3pt}
\end{table}

\section{Details on the Intrinsic Dimension Estimation.}
\label{sec:appendix_id}
Several recent works \cite{valeriani2023geometry, kvinge2023exploring, cheng2023bridging} have utilized ID to measure the complexity of latent representations in deep learning modeling. 
  Further \citet{pope2021the} has demonstrated a strong correlation between a dataset’s relative difficulty and its ID. 
We compute the ID of the latent representations using the TwoNN estimator \cite{facco2017estimating}, which relies solely on the distances between each point and its two nearest neighbors.
In practice, the \Th{TwoNN} estimator can be affected by noise, which
typically leads to an overestimation of the ID. Nevertheless, it is a robust tool to evaluate \emph{relative} complexity and has been used
effectively to analyze representations in deep neural networks \cite{valeriani2023geometry}. We adopt the \Th{TwoNN} implementation of \Th{DADApy}~\cite{glielmo2022dadapy}.

\section{Details on Evaluation Metrics}
\label{sec:appendix_metrics}

\subsection{Generation Metrics}
We follow the setup and use the same reference batches of ADM \cite{nichol21a}
for evaluation, utilizing their official implementation\footnote{\url{https://github.com/openai/guided-diffusion/tree/main/evaluations}}. We use NVIDIA A100 GPUs for our evaluation.
 We briefly explain each metric used for the evaluation.

 \begin{itemize}
     \item \textbf{FID} \cite{fid} quantifies the feature distance between the distributions of two image datasets by leveraging the Inception-v3 network \cite{szegedy2016rethinking}. The distance is calculated based on the assumption that both feature distributions follow multivariate Gaussian distributions.
     \item \textbf{sFID} \cite{sfid} computes FID with intermediate spatial features of the
        Inception-v3 network, to capture spatial distribution of the generated images
     \item \textbf{IS} \cite{is}  measures a KL-divergence between the original label distribution and the
distribution of Inception-v3 network's logits after the softmax normalization.
\item \textbf{Precision and Recall} \cite{kynkaanniemi2019improved} are the
fraction of realistic images and the fraction of training data covered by generated data respectivly.
 \end{itemize}

\clearpage

\subsection{Reconstruction Metrics}
We evaluate reconstruction on the validation set of Imagenet which contains 50K images. We provide a description of each metric used for the reconstruction evaluation.

 \begin{itemize}
     \item \textbf{PSNR} measures the quality of reconstructed images by comparing the maximum possible signal power to the level of noise introduced during reconstruction. Expressed in decibels (dB).
     \item \textbf{SSIM} \cite{ssim} assesses the similarity between two images by evaluating their structural information, luminance, and contrast. 
     \item \textbf{LPIPS} \cite{zhang2018unreasonable}  evaluates the perceptual similarity between two images by comparing their deep feature representations using VGG \cite{vgg}
 \end{itemize}

\subsection{Equivariance Error} 
\label{sec:equi_error}

To quantify the effectiveness of \texttt{EQ-VAE} at constraining the latent representations of the autoencoders to equivary under scale and rotation transformation we measure the equivariance error. Similar to \cite{Sosnovik2020Scale-Equivariant} we define the equivariance error as follows: \mbox{$\Delta_{eq}^{\mathcal{T}} = \frac{1}{\vert \mathcal{T}\vert \cdot N}  \underset{\vert \mathcal{T} \vert}{\sum} \underset{N}{\sum} \Vert \tau \circ \mathcal{E}(\mathbf{x}) - \mathcal{E}(\tau \circ  \mathbf{x}) \Vert_2^2 \;/ \; 
 \Vert \mathcal{E}(\tau \circ \mathbf{x})\Vert_2^2$  } where $\text{N}=50\text{K}$ in the number of samples in ImageNet validation and $\mathcal{T}$ is the set of transformations considered. We conduct our evaluation with $\mathcal{T}_r=  \{ \frac{\pi}{2},\pi, \frac{3\pi}{2}\}$ for rotations and $\mathcal{T}_s=  \{ 0.25, 0.50, 0.75\}$ for scale.

\section{Detailed Benchmarks}

\subsection{Detailed generative performance}
We provide a detailed evaluation of all the generative models presented in the main paper, including additional metrics and training iterations.  
 Specifically, ~\autoref{tab:details_dit_sit} details the performance of the \texttt{DiT-XL/2} and \texttt{SiT-XL/2} models, while ~\autoref{tab:detailed_repa} presents results for the \texttt{REPA} (\texttt{SiT-XL/2}) models trained with both \texttt{SD-VAE-FT-EMA} (as reported in the respective papers) and \texttt{EQ-VAE}. Additionally, ~\autoref{tab:detailed_maskgit} provides results for \texttt{MaskGIT} models trained using \texttt{VQ-GAN} and \texttt{EQ-VAE}. For all models, we use the evaluation metrics originally reported in the original publications.
\begin{table}[!h]
\centering
\setlength{\tabcolsep}{4pt}
\begin{tabular}{lccccccc}
\toprule
\Th{Model} & \Th{\#Iters.} &  \Th{FID$\downarrow$} & \Th{sFID$\downarrow$} & \Th{IS$\uparrow$} & \Th{Prec.$\uparrow$} & \Th{Rec.$\uparrow$} \\
\midrule

\texttt{DiT-XL/2} \cite{peebles2023scalable} & $400\text{K}$ & $19.5$ & $6.5$  & $77.5$ & $0.60$ & $0.60$ \\
w/ \our &  $50\text{K}$  & $73.6$ & $13.1$ \hspace{0.05cm} & $34.5$ & $0.50$ & $0.37$ \\
w/ \our &  $100\text{K}$ & $39.9$ & $6.8$  & $62.2$ & $0.60$ & $0.53$ \\
w/ \our &  $200\text{K}$ & $22.8$ & $5.9$  & $73.6$ & $0.61$ & $0.62$ \\
w/ \our &  $400\text{K}$ & $14.5$ & $5.6$  & $81.5$ & $0.63$ & $0.66$ \\
\midrule
\texttt{SiT-XL/2} \cite{ma2024sit} &  $400\text{K}$ & $17.2$ & $5.1$ & $76.5$ & $0.64$ & $0.63$ \\
w/ \our &  $50\text{K}$  & $76.1$ & $38.4$ \hspace{0.05cm} & $15.2$ & $0.50$ & $0.37$ \\
w/ \our &  $100\text{K}$ & $41.3$ & $10.9$ \hspace{0.05cm} & $30.9$ & $0.60$ & $0.53$ \\
w/ \our &  $200\text{K}$ & $24.9$ & $6.4$  & $54.6$ & $0.61$ & $0.62$ \\
w/ \our &  $400\text{K}$ & $16.1$ & $4.2$  & $79.7$ & $0.64$ & $0.66$ \\
\bottomrule
\end{tabular}
\vspace{-3pt}
\caption{\textbf{Detailed evaluation} for \texttt{DiT-XL/2} and \texttt{SiT-XL/2} models. All results are reported without classifier-free guidance ($\text{CFG}=1.0$).}
\label{tab:details_dit_sit}
\end{table}

\newpage
\begin{table}[!h]
\label{tab:diffbench}
\centering
\setlength{\tabcolsep}{4pt}
\begin{tabular}{cccccccc}
\toprule
\Th{Model} & \Th{\#Iters.} & \Th{FID$\downarrow$} & \Th{sFID$\downarrow$} & \Th{IS$\uparrow$} & \Th{Prec.$\uparrow$} & \Th{Rec.$\uparrow$} \\
\midrule 

\texttt{REPA} \cite{Yu2025repa} & $50\text{K}$ &  $52.3$ & $31.2$ \hspace{0.1cm} & $24.3$ & $0.45$ & $0.53$ \\
w/ \our  & $50\text{K}$ & $48.7$ & $26.3$ \hspace{0.1cm} & $27.6$ & $0.44$ & $0.53$ \\
\grayhline
\texttt{REPA} \cite{Yu2025repa} & $100\text{K}$ &  $19.4$ &  $6.1$ &  $67.4$ & $0.64$ & $0.610$ \\
w/ \our  & $100\text{K}$ & $18.7$ & $5.4$ & $67.8$ & $0.65$ & $0.59$ \\
\grayhline
\texttt{REPA} \cite{Yu2025repa} & $200\text{K}$ & $11.1$ & $5.1$ & $100.4$ \hspace{0.08cm} &  $0.69$ & $0.64$ \\
w/ \our  & $200\text{K}$ & $10.7$ & $5.1$ & $103.5$ \hspace{0.08cm} & $0.70$ & $0.62$ \\
\grayhline
\texttt{REPA} \cite{Yu2025repa} & $400\text{K}$ & $7.9$ \hspace{-0.26cm} & $5.1$ & $122.6$ \hspace{0.08cm} & $0.70$ &  $0.65$ \\
w/ \our  & $400\text{K}$ & $7.5$ \hspace{-0.26cm}  & $5.0$ & $128.8$ \hspace{0.08cm} & $0.71$ & $0.63$ \\

\bottomrule
\end{tabular}
\vspace{-3pt}
\caption{\textbf{Detailed evaluation} on \texttt{REPA} (\texttt{SiT-XL/2}) models. All results are reported without
classifier-free guidance ($\text{CFG}=1.0$)}
\label{tab:detailed_repa}
\vspace{-3pt}
\end{table}
\begin{table}[!h]
\centering
\setlength{\tabcolsep}{4pt}
\begin{tabular}{cccccc}
\toprule
\Th{Model} & \Th{Epochs} & \Th{FID$\downarrow$} & \Th{IS$\uparrow$} & \Th{Prec.$\uparrow$} & \Th{Rec.$\uparrow$} \\
\midrule 

\texttt{MaskGIT} \cite{chang2022maskgit} & $300$ &  $6.2$ &  $182.1$ &  $0.80$ & $0.51$ \\
\texttt{MaskGIT}* \cite{besnier2023pytorch} & $300$ & $6.8$ & $214.0$ & $0.82$ & $0.51$ \\

w/ \our  & $50$  \hspace{-0.26cm} & $11.1$ \hspace{0.08cm} & $116.1$ & $0.73$ & $0.52$ \\
w/ \our  & $100$ & $7.6$  & $167.5$ & $0.78$ & $0.53$ \\
w/ \our  & $200$ & $6.0$  & $211.8$ & $0.79$ & $0.55$ \\
w/ \our  & $300$ & $5.9$  & $228.7$ & $0.80$ & $0.55$ \\

\bottomrule
\end{tabular}
\vspace{-3pt}
\caption{\textbf{Detailed evaluation} on \texttt{MaskGIT} models. We use the open-source  \Th{PyTorch} reproduction for our experiments. All results are reported without
classifier-free guidance ($\text{CFG}=3.0$)}
\label{tab:detailed_maskgit}
\vspace{-3pt}
\end{table}

\subsection{Detailed reconstruction performance}
We detail in \autoref{tab:appendix_recon} the reconstruction evaluation metrics of each autoencoder with and without \texttt{EQ-VAE} regularization.
\label{sec:autoencoders_appendix}

\begin{table}[!h]

\centering
\setlength{\tabcolsep}{2.pt} 
\begin{tabular}{llcccc}
\toprule

& \Th{Autoencoder} & \Th{rFID$\downarrow$} & \Th{PSNR$\uparrow$} & \Th{LPIPS$\downarrow$} &  \Th{SSIM$\uparrow$} \\ 
\midrule 
 & \sdvae  & $0.90$ & $25.82$ & $0.146$ & $0.71$ \\
 & w/ \our (ours)   & $0.82$& $25.95$ & $0.141$ & $0.72$ \\ \addlinespace
\cline{2-6}
\addlinespace
\vspace{-10.0pt}
\multirow{6}{*}{\rotatebox{90}{\Th{Cont.}}} & & & & & \\

 & \sdxlvae  & $0.67$ & $27.36$ & $0.121$ & $0.76$ \\
 & w/ \our (ours)  & $0.65$ & $27.48$ & $0.118$ & $0.76$  \\ \addlinespace
\cline{2-6}
 \addlinespace

 & \texttt{SD3-VAE}  &  $0.20$  & $31.27$ & $0.060$ & $0.87$ \\
 & w/ \our (ours) &  $0.19$  & $31.06$ & $0.061$ & $0.87$ \\ 

\addlinespace
\cline{2-6}
 \addlinespace

 & \texttt{SD-VAE-16}  &  $0.87$  & $24.67$ & $0.161$ & $0.61$ \\
 & w/ \our (ours) &  $0.82$  & $25.21$ & $0.152$ & $0.69$ \\ 

 \bottomrule
 
 \addlinespace

\multirow{2}{*}{\rotatebox{90}{\Th{Disc.}}} & \texttt{VQ-GAN} & $7.94$ & $19.41$ & $0.540$ & $0.54$ \\
 & w/ \our (ours) & $7.54$ & $19.61$ & $0.510$ & $0.56$\\ \addlinespace
\bottomrule
\end{tabular}

\vspace{-3pt}
\caption{\textbf{Comparison of Autoencoders with and without
\texttt{EQ-VAE}}. Additional reconstruction evaluation metrics.}
\label{tab:appendix_recon}
\vspace{-3pt}
\end{table}

\clearpage

\section{Specifications of Autoencoder Models}
\label{sec:appendix:ae_specs}
\begin{table}[h]
\centering
\setlength{\tabcolsep}{4pt}
\begin{tabular}{ccccc}
\toprule
\Th{Autoencoder} & $\mathcal{L}_{reg}$ & \Th{Dataset} & $c$ & $f$  \\
\midrule 
 \texttt{SD-VAE} \cite{rombach2022high} & KL & OpenImages & 4 & 8 \\
 \texttt{SD-VAE-FT-EMA} \cite{rombach2022high} & KL & OpenImages + Laion Aesthetics & 4 & 8 \\
 \texttt{SD-VAE-16} \cite{rombach2022high} & KL & OpenImages & 16 & 16 \\
 \texttt{SDXL-VAE} \cite{podell2024sdxl} & KL & - & 4 & 8 \\
 \texttt{VQ-GAN} \cite{esser2021taming} & VQ & ImageNet & 256 & 16 \\
\bottomrule
\end{tabular}
\vspace{-3pt}
\caption{\textbf{Specifications of Autoencoders}. We provide additional information for the autoencoders used in our experiments regarding their original training dataset, latent channels $c$, and compression rate $f$.}
\label{tab:ae_specs}
\vspace{-3pt}
\end{table}

\section{Latent Generative Models}

Here we provide a brief description of the latent generative models, mentioned in the main paper:
\begin{itemize}[noitemsep, topsep=5pt]
    \item \texttt{MaskGIT} \cite{chang2022maskgit} utilizes a bidirectional transformer decoder to synthesize images by iteratively predicting masked visual tokens produced by a \texttt{VQ-GAN} \cite{esser2021taming}.
    
    \item \texttt{LDM} \cite{rombach2022high} proposes latent diffusion models, modeling the image distribution in a compressed latent space produced by a KL- or VQ-regularized autoencoder.

    \item \texttt{DiT} \cite{yao2024fasterdit} proposes a pure transformer backbone for training diffusion models and incorporates AdaIN-zero modules.

    \item \texttt{MaskDiT} \cite{zheng2023fast} trains diffusion transformers with an auxiliary mask reconstruction task.

    \item \texttt{SD-DiT} \cite{zhu2024sd} extends the \texttt{MaskDiT} architecture by incorporating a discrimination objective using a momentum encoder.

    \item \texttt{SiT} \cite{ma2024sit} improves diffusion transformer training by moving from discrete diffusion to continuous flow-based modeling.

    \item \texttt{REPA} \cite{Yu2025repa} aligns the representations of diffusion transformer models to the representations of self-supervised models.
\end{itemize}

\clearpage

\section{Additional Qualitative Results}

\begin{figure}[!h]
  \centering
    \resizebox{0.58\textwidth}{!}{  
    \setlength{\tabcolsep}{1.5pt}
    \begin{tabular}{cccc}
      \multirow{2}{*}{\small Input Image \textbf{x}}
      & \multicolumn{2}{c}{{\small \texttt{SD-VAE}}}
      & {\texttt{Ours}} \\
      \cmidrule(lr){2-3} \cmidrule(lr){4-4}
       
      & {\small$\mathcal{D} ( \mathcal{E}(\tau \circ \mathbf{x}))$}
      & {\small$\mathcal{D}(\tau \circ \mathcal{E}(\mathbf{x}))$}
      & {\small$\mathcal{D}(\tau \circ \mathcal{E}(\mathbf{x}))$} \\
      
      \vspace{-0.4cm} \\
      
     \includegraphics[width=0.15\linewidth]{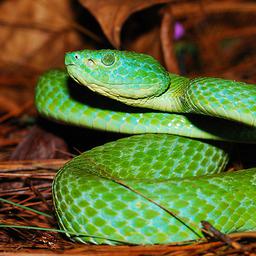} 
      & \includegraphics[width=0.15\linewidth]{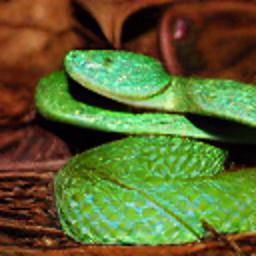} 
      & \includegraphics[width=0.15\linewidth]{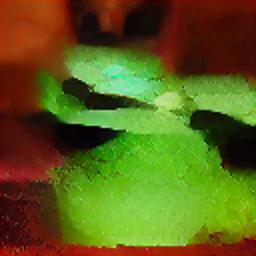} 
      & \includegraphics[width=0.15\linewidth]{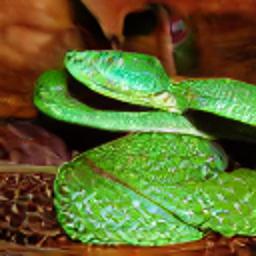} \\

      \includegraphics[width=0.15\linewidth]{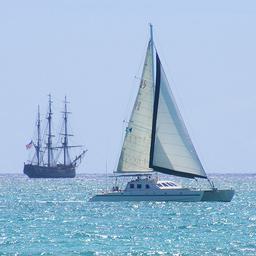} 
      & \includegraphics[width=0.15\linewidth]{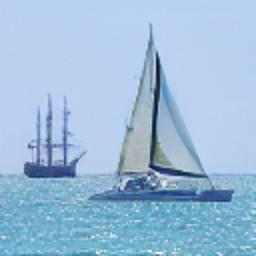} 
      & \includegraphics[width=0.15\linewidth]{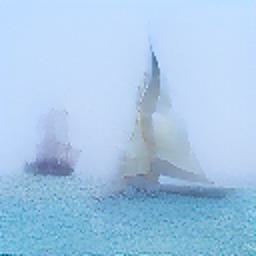} 
      & \includegraphics[width=0.15\linewidth]{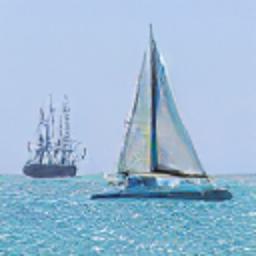} \\

      \mc{4}{(a) Scaling Transformation ($s=0.5$)}\\
      \vspace{-0.2cm} \\
      
      \includegraphics[width=0.15\linewidth]{fig/equi_plots/images/00000029.JPEG} 
      & \includegraphics[width=0.15\linewidth]{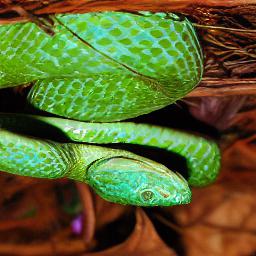} 
      & \includegraphics[width=0.15\linewidth]{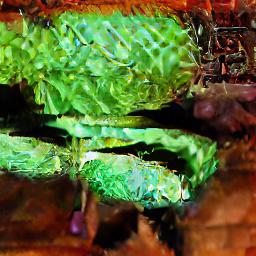} 
      & \includegraphics[width=0.15\linewidth]{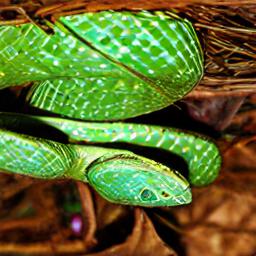} \\

          \includegraphics[width=0.15\linewidth]{fig/equi_latex_more/image_94.JPEG} 
      & \includegraphics[width=0.15\linewidth]{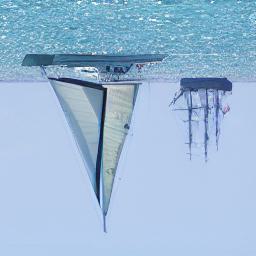} 
      & \includegraphics[width=0.15\linewidth]{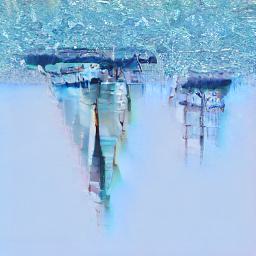} 
      & \includegraphics[width=0.15\linewidth]{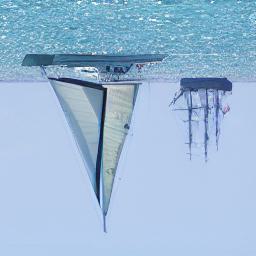} \\
      
      \mc{4}{(b) Rotation Transformation ($\theta=180^{\circ}$)
      }\\
    \end{tabular}
    }
  \caption{
  \textbf{Latent Space Equivariance.} 
  Reconstructed images using \texttt{SD-VAE}~\cite{rombach2022high} and \our when applying transformations $\tau$ to the input images ($\mathcal{D}(\mathcal{E}(\tau \circ \mathbf{x}))$) versus directly to the latent representations ($\mathcal{D}(\tau \circ \mathcal{E}(\mathbf{x}))$). We present results for scaling and rotation transformations $\tau$. Our approach preserves reconstruction quality under latent transformations, whereas \texttt{SD-VAE} exhibits significant degradation.
  }
  \label{fig:qualitative-equivariance-appendix}
\end{figure}

  \newcommand{\myfigA}[1]{\includegraphics[width=0.258\textwidth,valign=c]{#1}}


\begin{figure}[t]
    \centering
    \resizebox{0.8\textwidth}{!}{  
    \begin{tabular}{@{}c|c|c|c|c@{}}
        { \Large\texttt{\textbf{Image}}} &
        {\Large\texttt{\textbf{SD-VAE}}} &
        {\Large \makecell{\texttt{\textbf{+Ours}}}} &
        { \Large\texttt{\textbf{SDXL-VAE}}} &
        { \Large\makecell{\texttt{\textbf{+Ours}}}} \\
    
        \myfigA{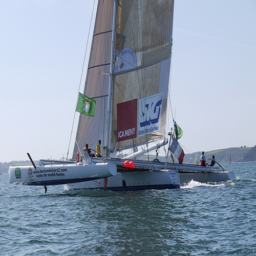} &
        \myfigA{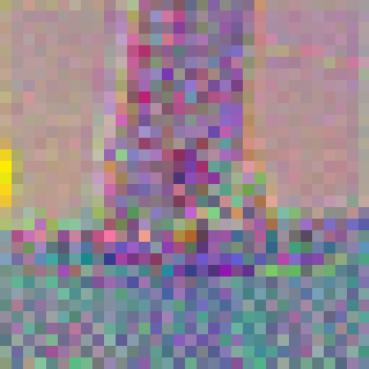} &
        \myfigA{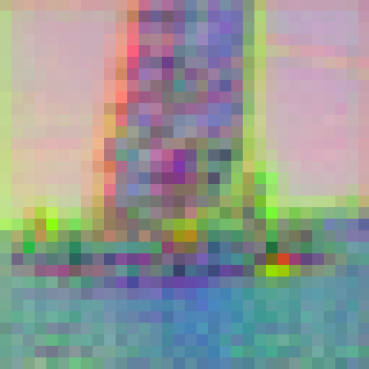} &
        \myfigA{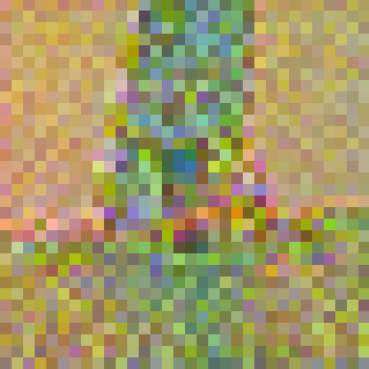} &
        \myfigA{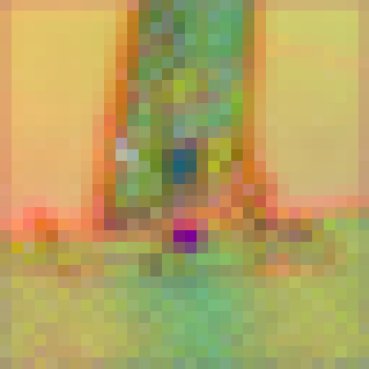} \\
    
        \multicolumn{5}{c}{\vspace{-2.1ex}}\\
    
          \myfigA{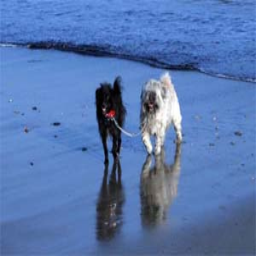} &
        \myfigA{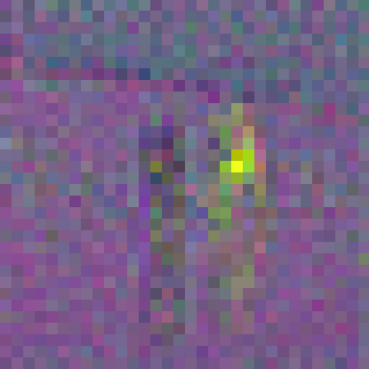} &
        \myfigA{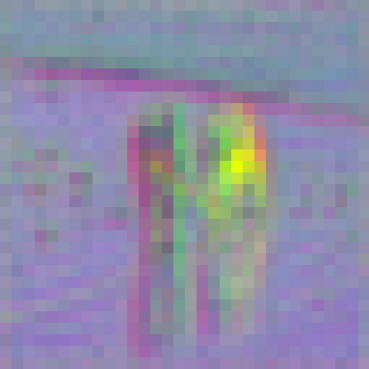} &
        \myfigA{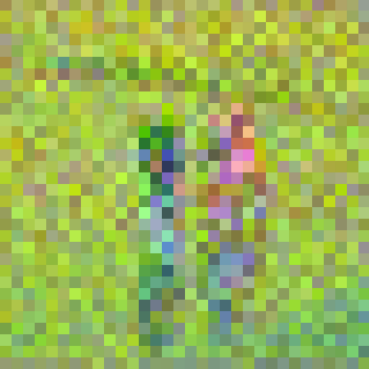} &
        \myfigA{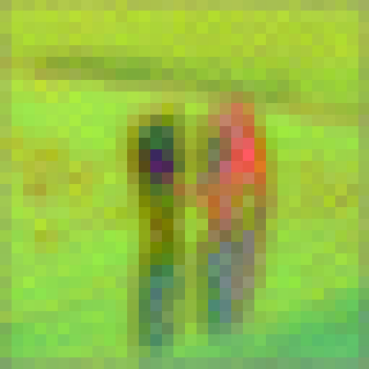} \\
        \multicolumn{5}{c}{\vspace{-2.1ex}}\\
    
          \myfigA{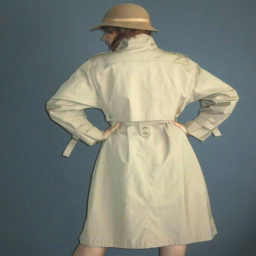} &
        \myfigA{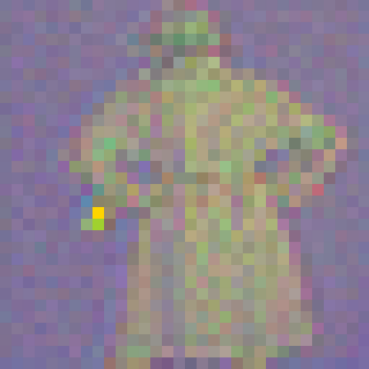} &
        \myfigA{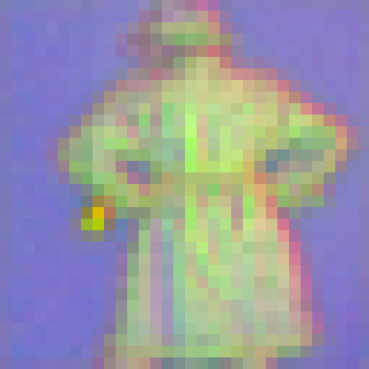} &
        \myfigA{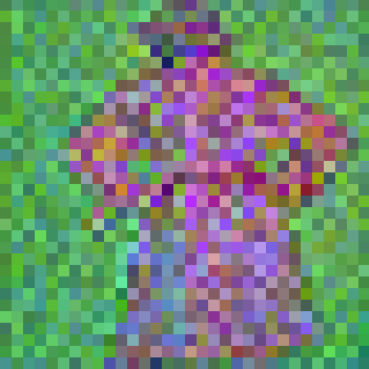} &
        \myfigA{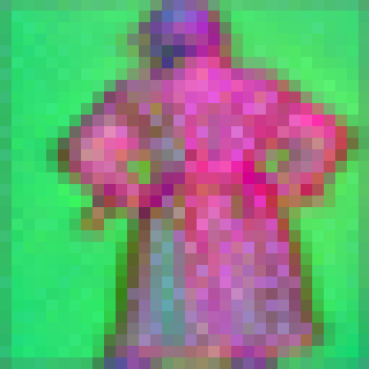} \\

        \multicolumn{5}{c}{\vspace{-2.1ex}}\\

          \myfigA{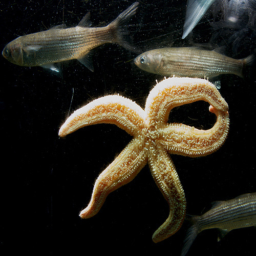} &
        \myfigA{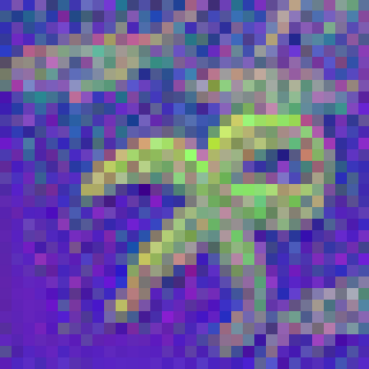} &
        \myfigA{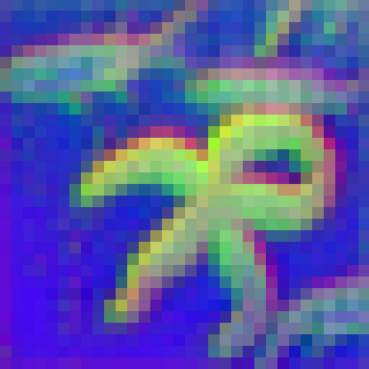} &
        \myfigA{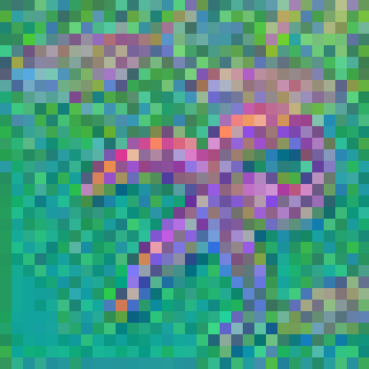} &
        \myfigA{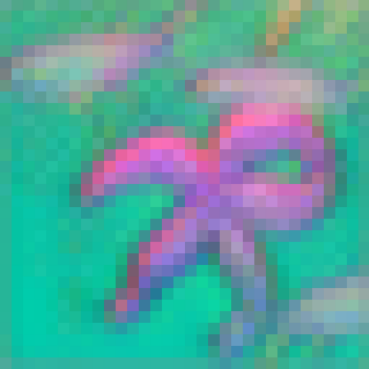} \\

        \multicolumn{5}{c}{\vspace{-2.1ex}}\\

          \myfigA{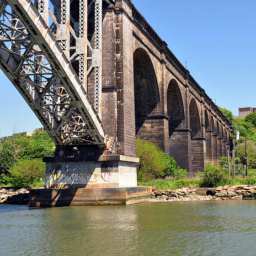} &
        \myfigA{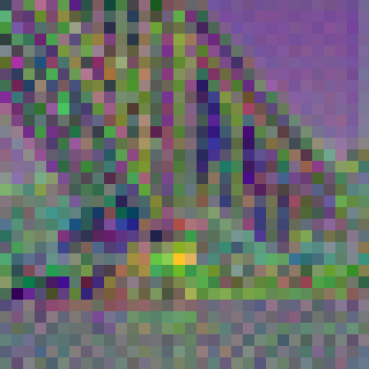} &
        \myfigA{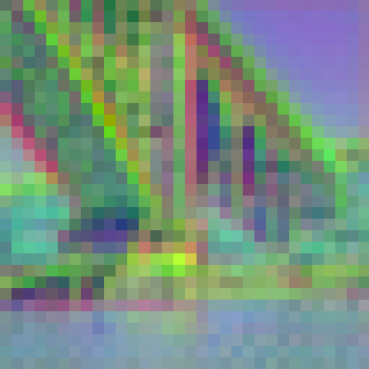} &
        \myfigA{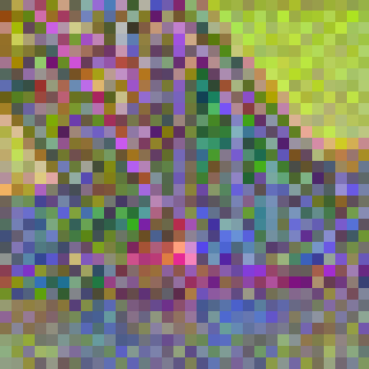} &
        \myfigA{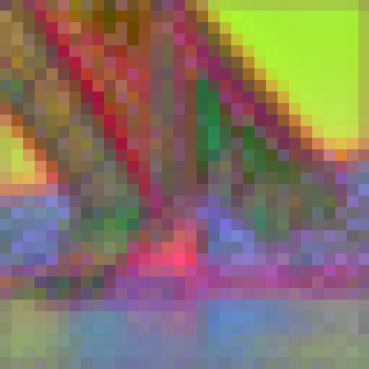} \\

        \multicolumn{5}{c}{\vspace{-2.1ex}}\\

          \myfigA{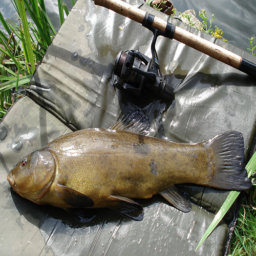} &
        \myfigA{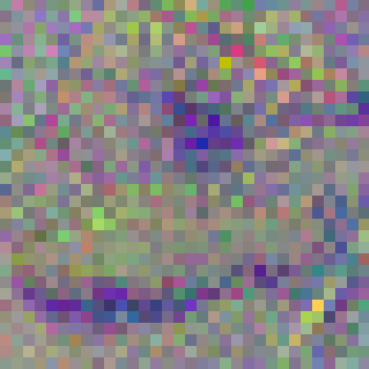} &
        \myfigA{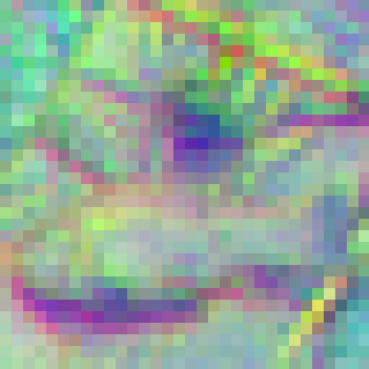} &
        \myfigA{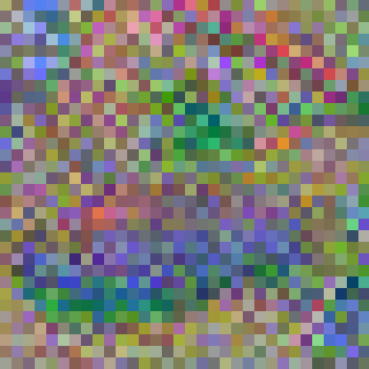} &
        \myfigA{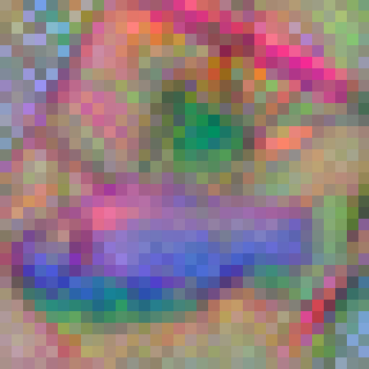} \\

    \end{tabular}
    }
    \caption{\textbf{Additional comparisons of latent representations} across different VAE models. \texttt{EQ-VAE} (+Ours) produces smoother latent representations for both \texttt{SD-VAE} and \texttt{SDXL-VAE}.}
    \label{fig:appendix_latents}
\end{figure}

\begin{figure}[t!]
    \vspace{-15pt}
    \centering
    {
\small
\centering
\newcommand{\resultsfignew}[1]{\includegraphics[width=0.09\textwidth,valign=t]{#1}}

\newcommand{\resultsfignewmedium}[1]{\includegraphics[width=0.184\textwidth,valign=t]{#1}}

\newcommand{\resultsfignewbig}[1]{\includegraphics[width=0.372\textwidth,valign=t]{#1}}
\setlength{\tabcolsep}{1pt}

\setlength{\tabcolsep}{1pt}

\begin{tabular}{@{}ccccccc@{}}  

\mc{7}{Class label = "panda" (388)}\\

\multicolumn{4}{c}{%
    \multirow{4}{*}{%
       \resultsfignewbig{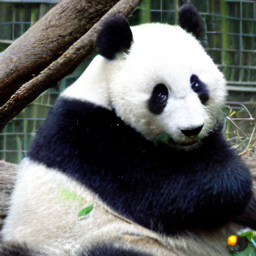}%
    }%
} & 

\multicolumn{2}{c}{%
    \multirow{2}{*}{%
       \resultsfignewmedium{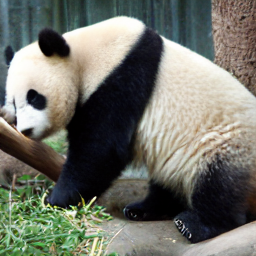}%
    }%
}
&
\resultsfignew{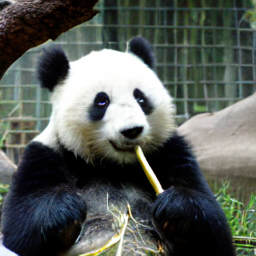}
\\
\mc{7}{\vspace{-2.0ex}}\\

& & & & 
& & 
\resultsfignew{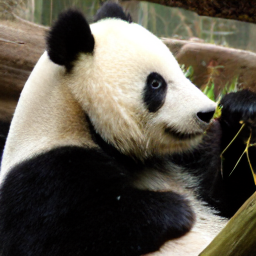} \\
\mc{7}{\vspace{-2.0ex}}\\

& & & & 
\multicolumn{2}{c}{%
    \multirow{2}{*}{%
       \resultsfignewmedium{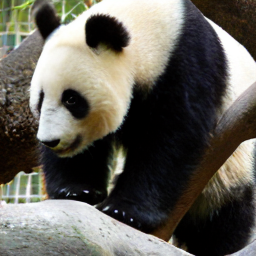}%
    }%
} & 
\resultsfignew{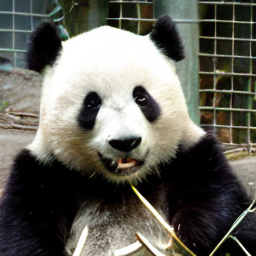} \\
\mc{7}{\vspace{-2.0ex}}\\

& & & & 
& & 
\resultsfignew{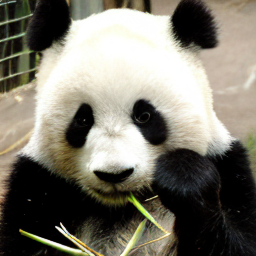} \\

\mc{7}{\vspace{-1.ex}}\\

\mc{7}{Class label = “golden retriever” (207)}\\

\multicolumn{4}{c}{%
    \multirow{4}{*}{%
       \resultsfignewbig{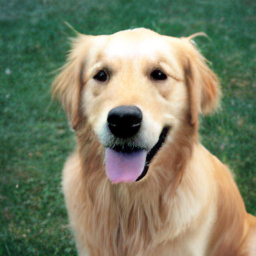}%
    }%
} & 

\multicolumn{2}{c}{%
    \multirow{2}{*}{%
       \resultsfignewmedium{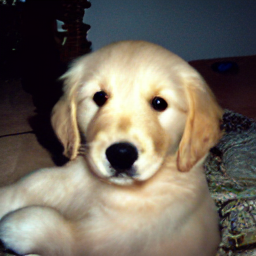}%
    }%
}
&
\resultsfignew{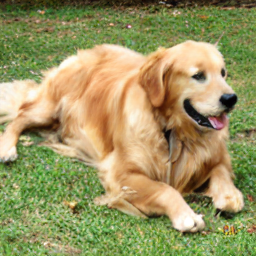}
\\
\mc{7}{\vspace{-2.0ex}}\\

& & & & 
& & 
\resultsfignew{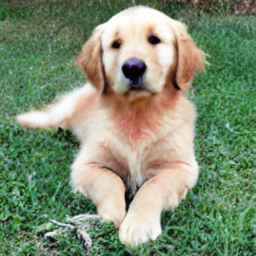} \\
\mc{7}{\vspace{-2.0ex}}\\

& & & & 
\multicolumn{2}{c}{%
    \multirow{2}{*}{%
       \resultsfignewmedium{fig/classes/207/000102.png}%
    }%
} & 
\resultsfignew{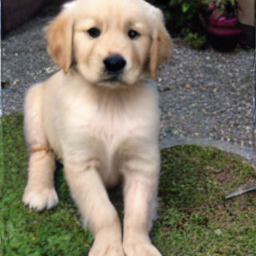} \\
\mc{7}{\vspace{-2.0ex}}\\

& & & & 
& & 
\resultsfignew{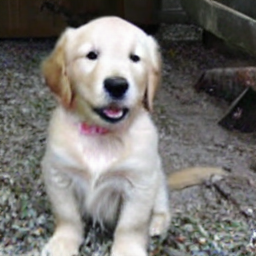} \\

\mc{7}{\vspace{-1.ex}}\\

\mc{7}{Class label = “macaw” (88)}\\

\multicolumn{4}{c}{%
    \multirow{4}{*}{%
       \resultsfignewbig{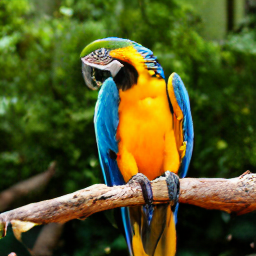}%
    }%
} & 

\multicolumn{2}{c}{%
    \multirow{2}{*}{%
       \resultsfignewmedium{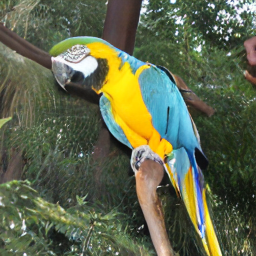}%
    }%
}
&
\resultsfignew{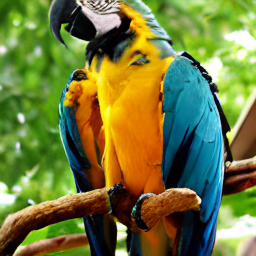}
\\
\mc{7}{\vspace{-2.0ex}}\\

& & & & 
& & 
\resultsfignew{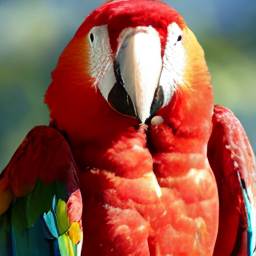} \\
\mc{7}{\vspace{-2.0ex}}\\

& & & & 
\multicolumn{2}{c}{%
    \multirow{2}{*}{%
       \resultsfignewmedium{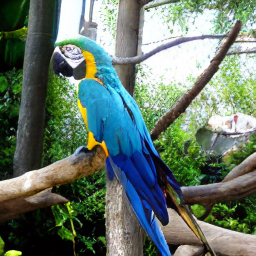}%
    }%
} & 
\resultsfignew{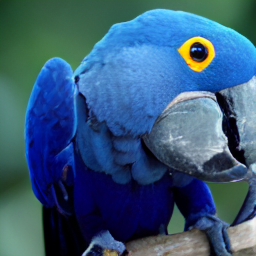} \\
\mc{7}{\vspace{-2.0ex}}\\

& & & & 
& & 
\resultsfignew{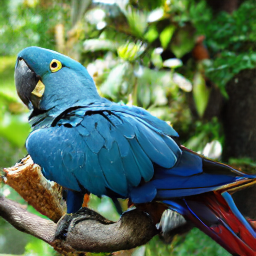} \\
    
\end{tabular}
}
    \vspace{-15pt}
    \caption{\textbf{Uncurated  samples $\mathbf{256 \times 256}$ \ditxltwo \texttt{/w EQ-VAE}.} Classifier-free guidance scale = 4.0.}
    \label{fig:appendix-gen-1}
\end{figure}

\end{document}